\documentclass[journal]{IEEEtai}

\usepackage[colorlinks,urlcolor=blue,linkcolor=blue,citecolor=blue]{hyperref}
\usepackage{pifont}
\usepackage{soul}
\usepackage{array}
\usepackage{graphicx}
\usepackage{ragged2e}
\usepackage{grffile}
\usepackage{cite}
\usepackage{booktabs}
\usepackage{amsmath,amssymb,amsfonts}
\usepackage{graphicx}
\usepackage{textcomp}
\usepackage[table,xcdraw]{xcolor}
\usepackage{placeins}
\usepackage{xcolor}
\usepackage{hyperref}
\usepackage{multirow}
\usepackage{url}
\usepackage{mwe}
\usepackage{algorithm}
\usepackage{algpseudocode}
\usepackage{subcaption}
\usepackage[leftcaption]{sidecap}

\def\IH#1{{\color{blue} {{#1}}}} 

\begin{document}

\title{COLT: Cyclic Overlapping Lottery Tickets for Faster Pruning of Convolutional Neural Networks}

\author{Md. Ismail Hossain, Mohammed Rakib, M. M. Lutfe Elahi, Nabeel Mohammed and Shafin Rahman

\thanks{Md. Ismail Hossain, Mohammed Rakib, M. M. Lutfe Elahi, Nabeel Mohammed, and Shafin Rahman are with the Department of Electrical and Computer Engineering at North South University, Dhaka 1229, Bangladesh. \textit{Corresponding author: Shafin Rahman} (e-mail: shafin.rahman@northsouth.edu)}
}

\markboth{Journal of IEEE Transactions on Artificial Intelligence, Vol. 00, No. 0, Month 2020}
{M. Hossain  \MakeLowercase{\textit{et al.}}: IEEE Journals of IEEE Transactions on Artificial Intelligence}

\maketitle

\begin{abstract}

Pruning refers to the elimination of trivial weights from neural networks. The sub-networks within an overparameterized model produced after pruning are often called Lottery tickets. This research aims to generate winning lottery tickets from a set of lottery tickets that can achieve accuracy similar to that of the original unpruned network. We introduce a novel winning ticket called Cyclic Overlapping Lottery Ticket (COLT) by data splitting and cyclic retraining of the pruned network from scratch. We apply a cyclic pruning algorithm that keeps only the overlapping weights of different pruned models trained on different data segments. Our results demonstrate that COLT can achieve similar accuracies (obtained by the unpruned model) while maintaining high sparsities. Based on object recognition and detection tasks, we show that the accuracy of COLT is on par with the winning tickets of the Lottery Ticket Hypothesis (LTH) and, at times, is better. Moreover, COLTs can be generated using fewer iterations than tickets generated by the popular Iterative Magnitude Pruning (IMP) method. In addition, we also notice that COLTs generated on large datasets can be transferred to small ones without compromising performance, demonstrating its generalizing capability. We conduct all our experiments on Cifar-10, Cifar-100, TinyImageNet, and ImageNet datasets and report superior performance than the state-of-the-art methods. The codes are available at: \texttt{\url{https://github.com/ismail31416/COLT}}
\end{abstract}

\begin{IEEEImpStatement}\label{impactstatement}
The emergence of large-scale deep learning models outperforms traditional systems in solving many real-life problems. Its success sometimes even beats human-level performance in several tasks. However, the main bottleneck is still the rise in computational cost. To minimize this, the researcher started looking for small-scale alternatives for large-scale models. One important direction is model pruning, which aims to prune the existing deep model without compromising performance. Existing approaches in this line of investigation perform iterative pruning of the same model in several pruning rounds. In this paper, we propose a novel idea to minimize the number of pruning rounds while keeping the unpruned model's accuracy. In other words, by utilizing our COLT algorithm, models can be pruned quicker, reducing the carbon footprint and making our algorithm more environment-friendly. Moreover, aligning with the literature, we demonstrate that the pruned subnetwork computed using one dataset can be used for different datasets without any dataset-specific pruning. It will help to build a pruned sub-network for a new domain quickly. This study will open up a new research prospect in finding a new pruning strategy for convolutional neural networks.
\end{IEEEImpStatement}

\begin{IEEEkeywords}
Model Compression, Pruning, Sparse Networks
\end{IEEEkeywords}

\section{Introduction}




\begin{figure*}[!t]
	\centering
    {\includegraphics[width=\linewidth]{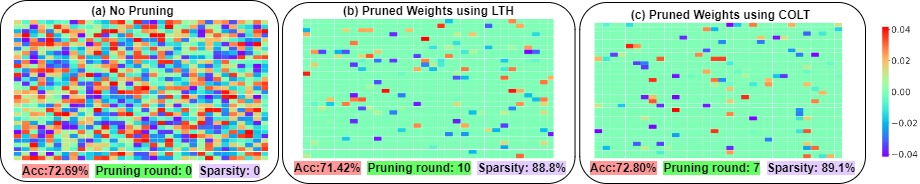}}
	\caption{We deal with the problem of pruning an overparameterized network without compromising performance. Here, we visualize the initial weights of learned networks while training with TinyImageNet. 
 (\textbf{a}) shows weights of the unpruned overparameterized network (no pruning), achieving an accuracy of 72.7\%. The popular pruning method, (\textbf{b}) LTH  can prune the same network to a sparsity of 88.8\% (\textcolor{green}{green} grid regions) in 10 pruning rounds, achieving an accuracy of 71.42\%. (\textbf{c}) In this paper, we propose a novel pruning method named COLT that can prune the network in \textbf{(a)} to a sparsity of 89.1\% in 7 pruning rounds and achieve an accuracy of 72.8\%. In comparison to LTH in \textbf{(b)}, COLT can generate highly sparse winning subnetworks (tickets) in fewer iterations (7 vs. 10), maintaining similar accuracy.}
	\label{fig:overviewDiagram}
\end{figure*}

Neural network pruning refers to removing unnecessary weights from neural networks. This problem has been widely studied since the 1980s to make networks sparse and efficient without adversely affecting performance \cite{r1,r2,r15,r16,r26,r27}. Consequently, pruning can compress the size of the model \cite{r15, r28} and make it less power-hungry \cite{r18, r29, r30}, possibly allowing inference to be more efficient. With the rise of deep neural networks (DNNs), pruning has witnessed a resurgence of interest. Many studies have been conducted towards making DNNs sparse in the image and vision computing field \cite{r13,r14,r15,r16,r17,r18,r19,r20,r40}. Pruning has consistently shown to be effective in reducing the high space and time demands of classification \cite{r37,r42} and object detection tasks \cite{r38,r41}. In this paper, we propose a novel neural network pruning strategy that can achieve higher compression, maintaining faster convergence and decent performance. 

A pruning technique, namely the lottery ticket hypothesis (LTH) \cite{r3}, has recently gained much attention from the research community. It is a weight-based pruning method, where the lowest magnitude weights are pruned iteratively after training. However, the aspect that differentiates LTH from other methods is resetting weights to their initial state (also called weight-rewinding) after pruning. Since multiple iterations of this train-prune-rewind cycle form the basis of this pruning algorithm, this process is called iterative magnitude pruning (IMP). According to \cite{r3}, a subnetwork (also called a ticket) generated using IMP can reach a sparsity of 90\% or more without compromising performance. So, once the sparse ticket is generated, it can be trained on the dataset from scratch and achieve accuracies equal to or greater than the original network. Later works like \cite{r5} demonstrate that these tickets generated from one dataset can even be transferred to another and achieve accuracies comparable to the original network. Specifically, they show that the subnetworks generated by IMP are independent of both datasets and optimizers. Considering such dataset transferability and optimizer independence of LTH tickets, we propose a new kind of lottery ticket that can achieve similar model compression (i.e., network pruning) but faster in computation (fewer pruning rounds) compared to LHT based method (see Fig.~\ref{fig:overviewDiagram}). 

Similar to IMP of the LTH-based pruning method, our pruning algorithm also utilizes IMP. We first split the training dataset into two class-wise partitions (Partitions 1 and 2). Each part contains non-overlapping classes and their instances. We train two models using Partition 1 and 2 independently but with the same initialization. Before the next iteration of the train-prune-rewind cycle of IMP, our pruning algorithm takes the intersection of the weights of the two models. In other words, only the overlapping weights survive, and non-overlapping ones become zero (i.e., pruned). This overlap is calculated by masks that find the subnetwork common in both models (trained on individual partitions). We notice that this intersection of the two lottery tickets is often a winning ticket. Since pruning occurs twice per iteration (from two separate models), our method can generate a sparse winning ticket in fewer iterations than LTH. Generally, IMP of LTH needs to run for several iterations before getting a sparse winning ticket. However, utilizing our iterative pruning algorithm, we can deduce a winning ticket of similar sparsity in fewer iterations than LTH. For our experiments, we have used ResNet-18 \cite{r32}, MobileNetV2 \cite{mobilenetv2}, and a 4-layer plain Convolutional Neural Network (CNN) called Conv-3 trained on Cifar-10 \cite{r25}, Cifar-100 \cite{r25}, Tiny ImageNet \cite{r3} and ImageNet\cite{r48}.
We have divided each of the three datasets into $N$ parts ($N$ = ${2,3,4}$), each having a different but equ`al number of classes. Our experimental results demonstrate that after finishing training-pruning-rewinding, if we only keep the overlapping weights between the two parts and prune the rest, it works as well as the original unpruned network for both datasets. This process is repeated multiple times until the desired sparsity for the network is reached. We call the pruned network a Cyclic Overlapping Lottery Ticket (COLT). COLT can be generated in fewer iterations than lottery tickets, which are generated from a single dataset. Moreover, COLT can also be used to initialize weights for another dataset with successful training outcomes. We have compared COLT's performance with LTH experimenting on Cifar-10, Cifar-100, Tiny ImageNet, and ImageNet datasets. We also have transferred COLT trained on Tiny ImageNet on Cifar-10 and Cifar-100. Beyond classification, we further assess the method on the Object Detection task. In experiments, we achieve performances similar to the original (unpruned network) network but take fewer pruning rounds than the LTH method. In summary, we make the following contributions: 

\begin{itemize}
\item We propose an iterative pruning algorithm for model compression based on overlapping/intersecting weights trained from $N$ non-overlapping partitions of the same dataset. The resulting subnetwork (after pruning) becomes high sparse and reaches the accuracy similar to that of the original unpruned network. We call this generated subnetwork/ticket Cyclic Overlapping Lottery Ticket (COLT).

\item Without compromising performance, COLT can achieve the desired sparsity by pruning a given network costing fewer iterations than the popular LTH-based pruning method.

\item We demonstrate the superiority of our method by experimenting on five datasets (Cifar-10, Cifar-100, Tiny ImageNet, ImageNet and Pascal-VOC) and three CNN architectures (Conv-3, ResNet-18, and MobileNetV2).

\end{itemize}


\section{Related works}\label{relatedworks}
\noindent\textbf{Traditional pruning:}\label{traditionalpruning} The overall goal of pruning is to reduce the network architecture by minimizing the number of network parameters without hampering the network performance. Han et al. \cite{r15} proposed the most popular pruning method, which suggests training the network to be pruned until convergence and then calculating a score for each parameter representing the importance of the given parameter. Later, parameters with low scores get pruned, resulting in network performance degradation. \cite{r15, r43} suggested finetuning the network using only the unpruned parameters. This train-prune cycle gradually makes the network sparse and is the traditional pruning method. Later, other related works proposed slight variations to this approach where weights are pruned while training in a periodic manner \cite{r34} or during initialization \cite{r35}. Another paper \cite{r17} explicitly adds extra parameters to the network to promote sparsity and serve as a basis for scoring the network after training. In this paper, instead of finetuning after pruning, we have rewound the unpruned weights to their initial state and then retrained them to convergence.

\begin{figure*}[!h]
	\centering
    	\subfloat{\includegraphics[width=0.48\linewidth]{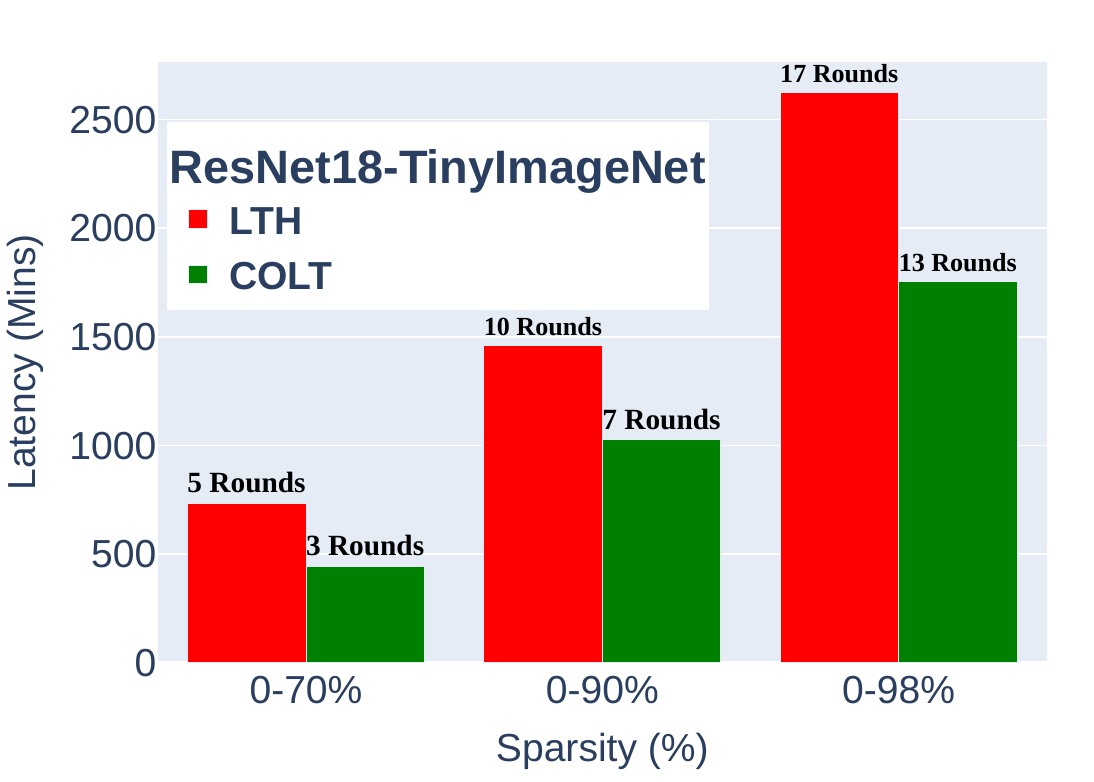} \label{mobilnetv2}}
    	\quad	    
    	\subfloat{\includegraphics[width=0.48\linewidth]{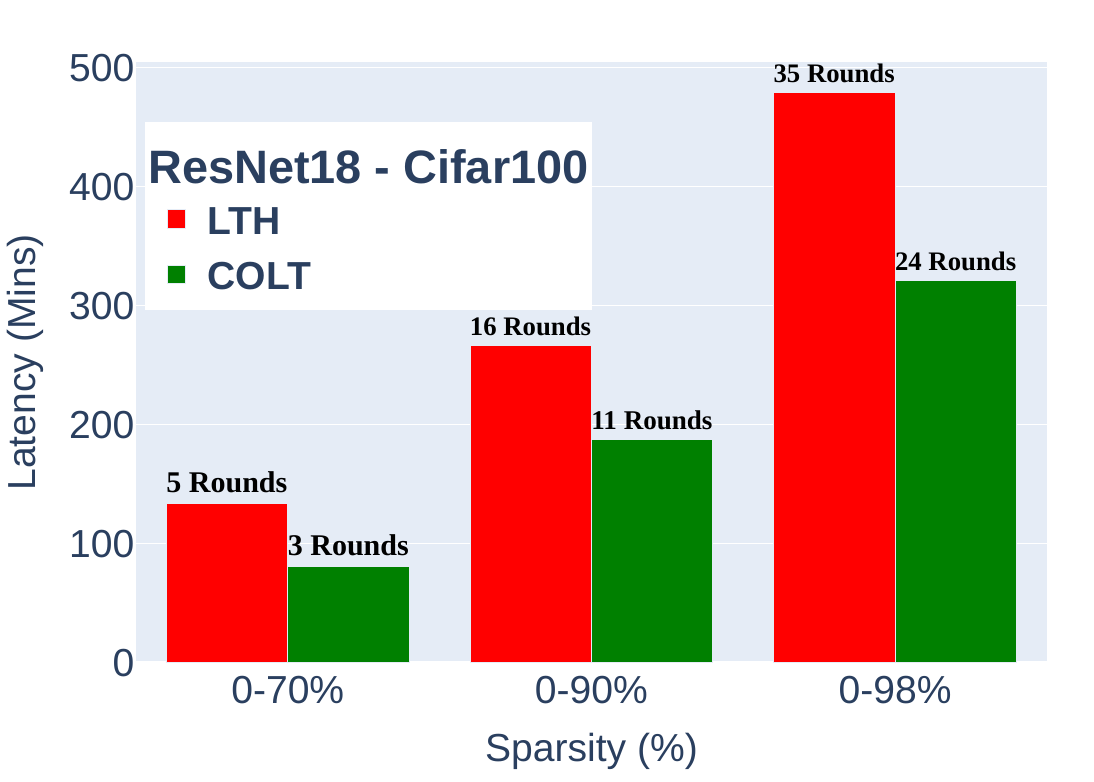} \label{resnet18}}
	\caption{ Illustration of the advantage of our proposed COLT over the LTH-based pruning method while using architecture ResNet-18 on (\textbf{Left}) Cifar10  and  (\textbf{Right}) Cifar-100 datasets. Compared to LTH, COLT can generate a highly sparse ticket in fewer rounds/iterations. We notice the same trend to achieve sparsity of different ranges, e.g.,  0-70\%, 0-90\%, and 0-98\%. In all cases, COLT maintains a similar accuracy to LTH while requiring fewer pruning rounds. Specifically, for (\textbf{Left}) ResNet-18 on Cifar-10, the sparsity 61.5\%, 88.8\% and 97.6\% achieved an accuracy of 60.28\%, 59.01\% and 51.81\% respectively. For (\textbf{Right}) ResNet-18 on Cifar100, the sparsity 61.7\%, 89.1\% and 97.4\% achieved an accuracy of about  73.59\%, 72.88\% and 68.53\%, respectively.}
	\label{fig:pruning-round}
\end{figure*}

\noindent\textbf{Pruning with lottery ticket hypothesis:}\label{pruningwithlotterytickethypothesis} In general, pruning is performed as a post-training step, necessitating training the full model before attempting parameter pruning. However, with the advent of the lottery ticket hypothesis (LTH) \cite{r3}, predetermined sparse networks can be trained from scratch. In contrary to previous works \cite{r7,r8,r9,r10,r11,r12}, LTH \cite{r3} suggested that over-parameterization is only required to find a ``good" initialization of a properly parameterized network. They demonstrated that substantially smaller sub-networks (lottery tickets) exist within large over-parameterized models. While trained in isolation, these sub-networks achieve similar or better performance than the original network even after pruning more than 90\% of the parameters. It allows networks to train with fewer resources and run inference of models on smaller devices like cellphones \cite{r4}. Also, it encourages generalization by acting as a regularizer for overparameterized models. To find and evaluate winning tickets using LTH, an overparameterized network is first initialized and trained to converge. Then the lowest magnitude weights are pruned, and the remaining weights are reset to their initial state at the start of training. This resetting of weights is also called weight rewinding. Finally, this smaller subnetwork (winning ticket) is then trained again to compare its performance with another subnetwork with the same number of parameters and the same initialization but randomly pruned weights. A good winning ticket outperforms a randomly pruned ticket. \cite{r44} provides a more rigorous definition of LTH by performing extensive experiments and exhaustively tuning hyperparameters like learning rate, and training epochs. In this paper, we propose an iterative pruning algorithm that uses fewer iterations to reach a specified sparsity.

\noindent\textbf{Pruning in transfer learning:}
Network pruning can take advantage of transfer learning concepts \cite{r21,r22,r23,r24}. Notable transfer learning-based methods \cite{r18, r20} train the model on one dataset, then apply pruning steps and later fine-tune the model on another dataset to achieve high performance. All of these papers analyze the transfer of learned representations of models. However, the transfer of weight initializations across datasets was first analyzed by \cite{r5}. It means transferring the initial weights after the train-prune-rewind cycle of IMP. \cite{r5} elaborately discussed the generalizing and transferring capability of lottery tickets. They demonstrate that winning ticket initializations can be generated independently of datasets and optimizers used in experiments. Furthermore, they also show that winning tickets generated by large datasets can successfully be transferred to small datasets. One could develop a robust ticket from a vast dataset and then reuse those initializations on other datasets without fully training a model. This paper demonstrates the transferable capability of tickets generated by COLT across various datasets. 

While all methods achieve pruning and high sparsity, COLT distinguishes itself by offering faster pruning and transferability, features not consistently present in other techniques. For instance, LTH  \cite{r3} emphasizes high sparsity alongside effective pruning but lacks the speed of pruning that COLT demonstrates. Similarly, SNIP  \cite{r35} and GraSP \cite{r10} achieve high sparsity, yet neither consistently matches our proposed COLT in terms of both speed and transferability across tasks.


\section{Methodology}\label{methodologysection}

\subsection{Problem formulation} 
Let us consider a feed-forward network $\mathcal{F}(x;\theta)$ with initial parameters $\theta = \theta_0 \sim \mathcal{D}_\theta$ and input $x$. After training until convergence, this network reaches an accuracy of $a$ achieving a minimum validation loss. $\mathcal{F}$ can also be trained with a mask $m \in \{0,1\}^{|\theta|} $ operated on parameters by $m \odot \theta$ achieving a sparsity $s\%$. To generate this mask $m$, we need to perform $j$ rounds of train-prune-rewind cycle.  Our goal is to design such a mask $m$ that when operated on parameters, $\theta$ (by $m \odot \theta$) generates a network, $\mathcal{F}(x; m \odot\theta)$ achieving a commensurate accuracy, $a' \geq a$ using less training time and pruning rounds $j' \leq j$ while maintaining the same sparsity $s\%$. Training with $m = 1^{|\theta|}$ means we train an unpruned network $\mathcal{F}$. While using $m \in \{0,1\}^{|\theta|} $, we train a pruned subnetwork of $\mathcal{F}$. In this case, the corresponding weights/parameters inside $\theta$, where $0$s are assigned inside $m$ have been pruned. More zeros inside $m$ creates more sparsity inside network parameters when applying $m \odot \theta$. We aim to increase such sparsity in $\mathcal{F}$ without compromising performance and training time.

\noindent\textbf{Lottery ticket hypothesis (LTH):} Frankle and Carbin \cite{r3} proposed the most popular solution for this problem. Later \cite{r5,zhou2019deconstructing} improved the LTH idea. LTH used a magnitude-based pruning algorithm. By identifying a sparse mask, $m$, lower magnitude weights are removed in different pruning rounds. The remaining weights are then rewound to their initial state by $\theta = m \odot \theta$. This rewound set of weights is a lottery ticket. If one can find a lottery ticket that can train a model and achieve accuracy equal to or greater than the initial unpruned weights, that lottery ticket is called the winning ticket. The train-prune-rewind cycle is performed multiple times iteratively to generate highly sparse winning tickets.

\noindent\textbf{Issues with LTH:} Repetitively training from scratch to the convergence of large networks while finding a winning ticket is a costly process, increasing computational time. In Fig. \ref{fig:pruning-round}, we see that to reach a highly sparse ticket, LTH requires a high no. of pruning rounds (i.e., increasing latency). For example, it takes 62 pruning rounds to generate a 98\% sparse ticket of MobileNetV2 on Cifar-100. In this paper, we aim to reduce the number of pruning rounds (training time) to reach such a high sparsity.

\subsection{Solution strategy}\label{solutionstrategy}
Morcos et al. \cite{r5} observed an intriguing fact about LTH regarding the transferability of lottery tickets. Pruned subnetworks, i.e., winning tickets computed using one dataset, can be transferred to other datasets, given the domain of data is similar (e.g., natural images). They argue that the same pruned subnetworks can generalize across different training conditions, optimizers, and datasets. The main implication of this finding is that we can compute a generic winning ticket once using one dataset and then reuse it across multiple datasets or optimizers. It can save the cost of repeatedly computing optimizers or dataset-specific tickets. The winning ticket can achieve similar accuracy and sparsity identical to the winning ticket of the entire model LTH training on the novel dataset. In this paper, we investigate the transferability property of LTH. We observe that intermediate tickets (before obtaining the winning tickets) also have the transferability property but with less sparsity. Each pruning round of our pruning method produces some intermediate subnetworks/tickets, perform the intersection of different subnetworks/tickets, and then transfer the new ticket to the next round. Our concept of transferring tickets across iteration/pruning rounds helps us achieve a similar performance of LTH, costing fewer rounds of computation.


\begin{table}
\centering
\caption{Transferring LTH tickets calculated from Partition 1 and 2 to the entire Cifar-100 dataset. The performance of overlapping tickets achieves better sparsity without compromising accuracy. It confirms that overlapping also has transferability property in addition to individual partitions' tickets.}
\label{tab:motivation}
\begin{tabular}{cccc} 
\hline
Using tickets from & Partition-1 & Partition-2 & Overlapping ticket \\ 
\hline
Accuracy (\%)      & 72.4     & 72.3     & 69.1        \\ 
Sparsity (\%) & 79.9    & 79.9     & 95.0        \\
\hline
\end{tabular}
\end{table}

Our motivation begins with experiment results shown in Table \ref{tab:motivation}. We divide Cifar-100 into $N$ partitions. Usually, N is used as a hyperparameter. In the case of $N$=2, if we have 100 classes for a dataset, then images of the first 50 classes remain in one partition and the remaining 50 in another. Then we generate LTH tickets (until around 80\% sparsity) from each partition. According to \cite{r5}, both tickets from two separate partitions are transferable to the entire  Cifar-100 dataset. Because of this, we achieve 72.4/79.9\% and  72.3/79.9\% accuracy/sparsity based on training from partitions 1 and 2, respectively. We know there exists a strong correlation between tickets generated by both partitions. It motivates us to check the performance of overlapping tickets, meaning the intersecting sub-network, which remains common after pruning. We notice that even excluding non-overlapping weights (increasing sparsity to 95\%), we can still achieve respectable performance (69.1\%) on full Cifar-100. It tells that in addition to improving sparsity, overlapping tickets calculated from multiple partitions of the same dataset are also transferrable to the entire dataset. Since intermediate tickets obtained from the intermediate LTH pruning rounds are transferable, we calculate the overlapping ticket of N different partitions in each pruning round. It can provide us with further pruning, which leads us to fewer pruning rounds than LTH, maintaining transferability to individual partitions and the full dataset. In each pruning round, we train $N$ models based on $N$ partitions of the same dataset to calculate $N$ tickets from each partition, then calculate the overlapping tickets between them, and finally, transfer the overlapping ticket to the next pruning round for initialization of those $N$ models. We continue this until the overlapping ticket achieves similar accuracy to the unpruned model.

\begin{figure}[!t]
	\centering
	\vspace{-1em}
    {\includegraphics[width=1\linewidth]{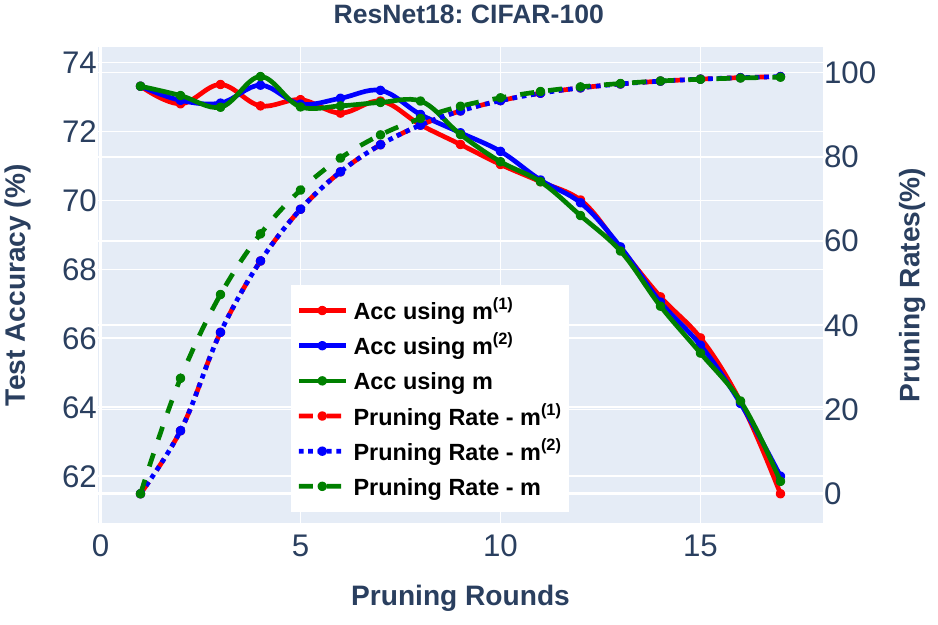}}
    \vspace{-1em}
	\caption{Transferability of tickets calculated from Partition 1, $m^{(1)}$, partition 2, $m^{(2)}$ and their overlap, $m = m^{(1)} \cap m^{(2)}$. Overlapping ticket, $m$ achieves a higher pruning rate (sparsity), maintaining similar accuracy to others. $m^{(1)}$ and $m^{(2)}$ get matching sparsity because we prune fixed $p\%$ low magnitude weight in every pruning round. In later pruning rounds, the behavior of all tickets is similar because every time the same $m$ calculated from the current round is transferred to both $\mathcal{F}_1$ and $\mathcal{F}_2$ for the next round making $\theta^{(1)}$ and $\theta^{(2)}$ similar.}
	\label{fig:roundwise}
\end{figure}

\sidecaptionvpos{table}{c}
\begin{SCtable*}
\centering
\caption{A toy example. Initial random weights are updated after training. Using updated weights, we calculate $m^{(1)}$ and $m^{(2)}$ by assigning zero to the locations where (\textbf{bold} values) $p\%$ lower magnitude weights exist. Then, we calculate a generic ticket, $m = m^{(1)} \cap m^{(2)}$, that refers to the pruned subnetwork. Pruned initial weights for the subsequent pruning round are calculated by $m\odot\theta^{(1)}$ and $m\odot\theta^{(2)}$.}
\label{tab:toyexample}
\scalebox{.7}{
\begin{tabular}{lllcl} 
\multicolumn{1}{c}{Initial weights}                                                                      & \multicolumn{1}{c}{After training}                                                                                         & \multicolumn{1}{c}{Mask}                                                            &  Overlapping mask                                                  & \multicolumn{1}{c}{Pruned weights}                                                                 \\ 
$\theta^{(1)}=$                                                                                         &  $\theta^{(1)}=$                                                                                                                          & $m^{(1)}=$                                                                          & \multicolumn{1}{l}{$m = m^{(1)} \cap m^{(2)}$}                                & $m\odot\theta^{(1)}=$                                                                                                   \\
\multicolumn{1}{c}{$\begin{bmatrix}0.1 & -0.2 & 0.9 \\-0.4 & 0.6 & 0.8 \\0.3 & 0.5 & -0.7\end{bmatrix}$} & \multicolumn{1}{c}{$\begin{bmatrix}0.8 & -0.3 & 0.4 \\\textbf{-0.1} & \textbf{0.2} & 0.7 \\0.9 & 0.5 & -0.5\end{bmatrix}$} & \multicolumn{1}{c}{$\begin{bmatrix}1 & 1 & 1 \\0 & 0 & 1 \\1 & 1 & 1\end{bmatrix}$} & \multirow{4}{*}{$\begin{bmatrix}1 & 1 & 0 \\0 & 0 & 1 \\1 & 1 & 1\end{bmatrix}$} & \multicolumn{1}{c}{$\begin{bmatrix}0.1 & -0.2 & 0 \\0 & 0 & 0.8 \\0.3 & 0.5 & -0.7\end{bmatrix}$}  \\
                                                                                                         &                                                                                                                            &                                                                                     &                                                                                  &                                                                                                    \\
$\theta^{(2)}=$                                                                                           &  $\theta^{(2)}= $                                                                                                                         & $m^{(2)}=$                                                                           &                                                                                  &$m\odot\theta^{(2)}=$                                                                                                     \\
\multicolumn{1}{c}{$\begin{bmatrix}0.1 & -0.2 & 0.9\\-0.4 & 0.6 & 0.8 \\0.3 & 0.5 & -0.7\end{bmatrix}$}  & \multicolumn{1}{c}{$\begin{bmatrix}0.9 & -0.7 & \textbf{0.2} \\\textbf{-0.1} & 0.4 & 0.5\\0.8 & 0.3 & -0.6\end{bmatrix}$}  & \multicolumn{1}{c}{$\begin{bmatrix}1 & 1 & 0 \\0 & 1 & 1 \\1 & 1 & 1\end{bmatrix}$} &                                                                                  & \multicolumn{1}{c}{$\begin{bmatrix}0.1 & -0.2 & 0 \\0 & 0 & 0.8 \\0.3 & 0.5 & -0.7\end{bmatrix}$}  \\
\end{tabular}
}
\end{SCtable*}
\subsection{Cyclic Overlapping Lottery Ticket (COLT)} \label{pruningalgorithmsection}

Let $\mathcal{Y}$ be the set of classes in our training dataset $D = \left \{{x}_{i},{y}_{i} \right \}_{i=1}^n$, where, $x_i \in \mathbb{R}^{h\times w}$ and ${y}_{i} \in \mathcal{Y}$  represent the input image and class label, respectively, and $n$ denotes the number of instances in the dataset. $\mathcal{F}(x;\theta)$ is the unpruned model with all parameters $\theta$. We aim to calculate a winning ticket represented as a sparse mask, $m \in \{0,1\}^{|\theta|}$, such that $\mathcal{F}(x;m\odot\theta)$ can achieve similar accuracy of the unpruned model. We propose a novel approach for generating winning tickets that can be calculated using fewer pruning rounds than LTH.

Let initialize a neural network $F(x; \theta_0)$, where the initial parameters $\theta = \theta_0$.  Subsequently, the dataset $D$ is partitioned into two disjoint subsets $D^{(1)}$ and $D^{(2)}$, ensuring $D = D^{(1)} \cup D^{(2)}$ and $D^{(1)} \cap D^{(2)} = \emptyset$. Two models, $F_1$ and $F_2$, are then trained to convergence on $D^{(1)}$ and $D^{(2)}$, respectively, resulting in trained parameters $\theta_1$ and $\theta_2$ with accuracies $a_1$ and $a_2$.
The core novelty of this method lies in cyclic pruning with overlapping tickets. During the pruning process, the model's parameters $\theta_0$ are iteratively masked by $N$ subnetworks represented by masks $m_i \in {0, 1}^{|\theta_0|}$. Each mask $m_i$ removes the lowest $p\%$ magnitude weights. Pruned parameters $\theta_i$ for each mask are obtained using element-wise multiplication $\theta_i = m_i \odot \theta$. These masks collectively form the combined mask $M = \cap_{i=1}^N m_i$, ensuring an overall sparsity level $s$ while maintaining or improving performance. After obtaining the combined mask $M$, it is applied to the model's parameters $F(x; M \odot \theta)$.

The method consists of four key steps described below:

\noindent\textbf{(a) Data partitioning:} We partition the training data into $N$ non-overlapping halves based on class labels. For this, we randomly split the set of classes $\mathcal{Y}$ into $N$ equal non-overlapping halves $\mathcal{Y}^{(1)}$, $\mathcal{Y}^{(2)}$...$\mathcal{Y}^{(N)}$, where, $\mathcal{Y}^{(1)}\cap \mathcal{Y}^{(2)}\cap \mathcal{Y}^{(N)}= \emptyset$. For an example of the class-specific partition of two sets , we create two separate datasets:
$D^{(1)} = \left \{{x}^{(1)}_{i},{y}^{(1)}_{i} \right \}_{i=1}^{n^{(1)}}$ and
$D^{(2)} = \left \{{x}^{(2)}_{i},{y}^{(2)}_{i} \right \}_{i=1}^{n^{(2)}}$
where, ${y}^{(1)}_{i} \in \mathcal{Y}^{(1)}$ and ${y}^{(2)}_{i} \in \mathcal{Y}^{(2)}$. We prefer the class-based split because it can provide us with more diversity inside intermediate tickets (compared to instance-based alternatives), which will help in faster pruning. For the main experiment (in Fig. \ref{results}), we split the dataset into only two halves because each half will contain enough instances for generalization. More than two partitions (three/fours) may end up with smaller subsets of the dataset. The effect of multiple partitions is further discussed in Sec. \ref{multiplepartitions}

\noindent\textbf{(b) Model training:} Using the partitioned data $D_1$ and $D_2$, we train two separate models $\mathcal{F}_1(x^{(1)};m\odot\theta^{(1)})$
and $\mathcal{F}_2(x^{(2)};m\odot\theta^{(2)})$, respectively. At the first pruning round, $m = 1^{|\theta|}$ meaning training of the unpruned model. In later pruning rounds, $m$ will be determined by the mask generation process discussed in the next paragraph, where $m$ will become sparse, containing zeros in the pruned locations. After calculating mask, $m$, we apply the same mask on randomly initialized parameters by $m\odot\theta^{(1)}$ and $m\odot\theta^{(2)}$. The $\odot$ operation is actually performing the pruning because it brings sparsity in weight matrices. At the beginning of the training, parameters $\theta^{(1)}$ and $\theta^{(2)}$ will get exactly the same initial weights as $\theta_0 \sim \mathcal{D}_\theta$ that will later be updated during the course of backpropagation. Therefore, after convergence of models, $\mathcal{F}_1$ and $\mathcal{F}_2$, both $\theta^{(1)}$ and $\theta^{(2)}$ are expected to contain different weights. A toy example of initial and after training weight of $\theta^{(1)}$ and $\theta^{(2)}$ are shown in Table \ref{tab:toyexample}. Note that training of $\mathcal{F}_1$ and $\mathcal{F}_2$ are independent, meaning a parallel implementation is possible. Moreover, associated datasets $D_1$ and $D_2$ are smaller than the original ones. It means that we can minimize the overall training time.

\noindent\textbf{(c) Mask generation:} Based on the trained weights $\theta^{(1)}$ and $\theta^{(2)}$ from two separate model, we calculate a mask, $m$. First, we generate two intermediate masks, $m^{(1)}$ and $m^{(2)}$ from $\theta^{(1)}$ and $\theta^{(2)}$, respectively. Analyzing the weights of $\theta^{(1)}$, we prune $p\%$ of lower magnitude weights. Thus, the binary mask $m^{(1)}$ will get zeros and ones where weights need to be pruned and not pruned, respectively. Similarly, we generate $m^{(2)}$ from $\theta^{(2)}$. In our experiment, we use $p = 20$ as used in LTH \cite{r3}. Note we use a small $p$ because a large $p$ gains more sparsity but, at the same time, increases the chance of layer collapse. Now, we generate $m = m^{(1)} \cap m^{(2)}$ to identify overlapping weight locations. $m$ describes the common tickets where both models agree. Our method achieves faster pruning because of the pruning from $p\%$ low-magnitude weights and intersecting masks ($m$) calculated from two different partitions of the dataset. In Fig. \ref{fig:roundwise}, we show the performance of $m^{(1)}$, $m^{(2)}$ and $m$ on entire Cifar-100. It shows further evidence that the overlapping ticket, $m$, is transferable across different pruning rounds, ensuring higher sparsity.

\noindent\textbf{(d) Rewinding:} In this step, we are transferring the tickets calculated from both partitions to individual partitions of the dataset. Using the calculated $m$ from the previous step, we repeat the process for the subsequent pruning round by iterating to step \textbf{(b)}. The same $m$ will modify $\theta^{(1)}$ and $\theta^{(2)}$. Models' training rewinds the so far unpruned weights (after $\odot$ operation) to their initial values used during the initialization at the beginning of the training. Using the $m$ calculated from the previous round facilitates training such that our method tries to prune within so far unpruned weights only in the next iteration rather than considering all parameters.

\begin{figure}[!t]
	\centering
	\vspace{-1em}
    {\includegraphics[width=1\linewidth]{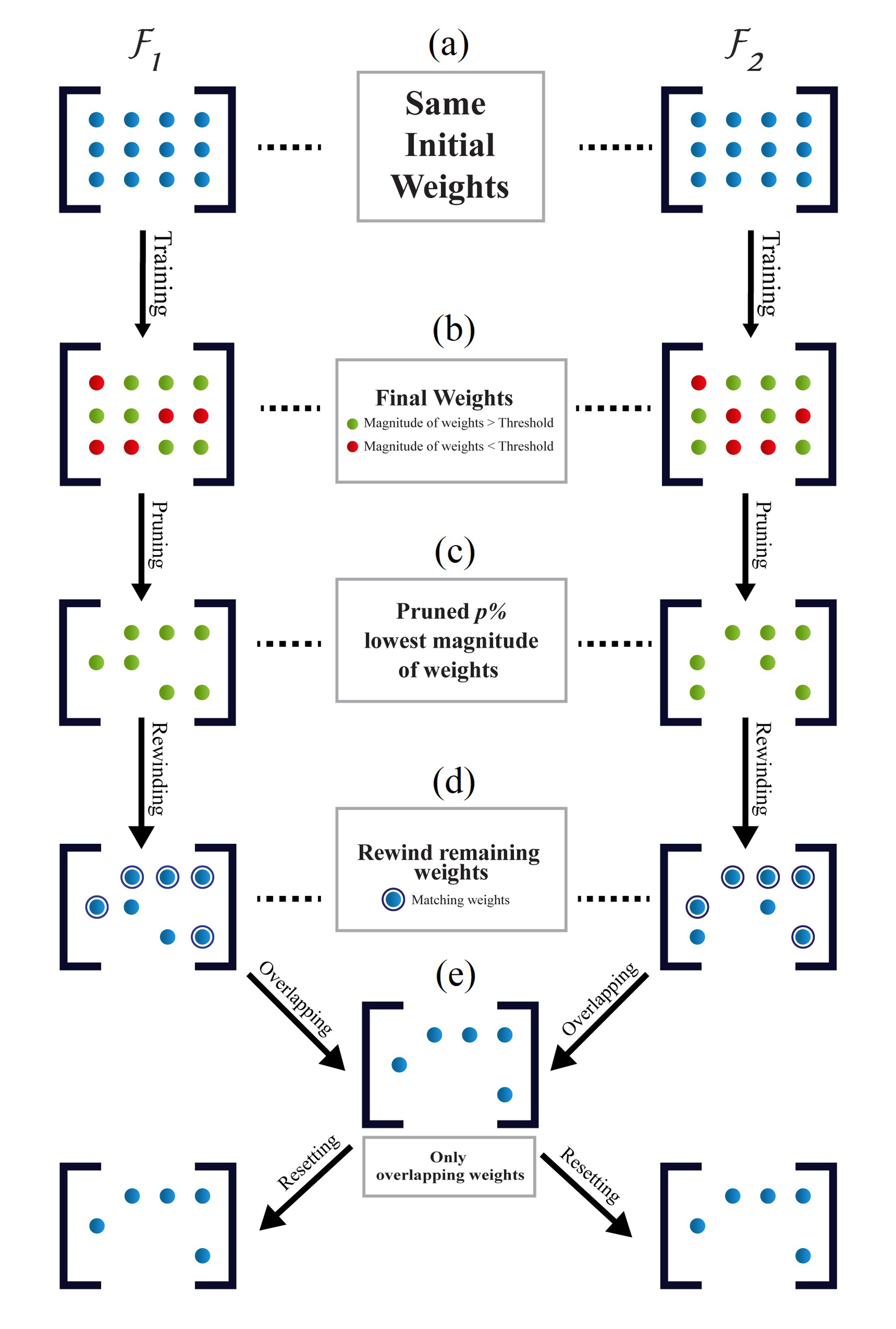}}
    \vspace{-2em}
	\caption{Visual illustration of COLT generation while $N$=2. \textbf{(a)} Two models, $\mathcal{F}_1$ \& $\mathcal{F}_2$, of identical architectures, are first initialized with the same initial weights (\textcolor{blue}{blue}). \textbf{(b)} Next, the two models are trained using the same set of hyperparameters to deduce the final weights (\textcolor{red}{red} \& \textcolor{green}{green}). \textbf{(c)} After training, p\% lowest magnitude final weights are pruned from each model. (\textcolor{red}{red} weights are the p\% lowest magnitude weights while the \textcolor{green}{green} ones are above the threshold) \textbf{(d)} Then the models are rewound to their initial state of weights (\textcolor{blue}{blue}, with the pruned weights being zero. \textbf{(e)} After rewinding, the models' overlapping weights (\textcolor{blue}{blue}) are kept, and the rest are pruned. The result is a COLT ticket that can gain accuracies similar to or better than the original network.}
	\label{fig:colt_illustration}
\end{figure}

\begin{algorithm}[!t]
\caption{Cyclic Overlapping Lottery Ticket (COLT)}
\label{algorithm}
\begin{algorithmic}[1]
\Require Dataset $D$, initial network parameters $\theta$, number of pruning rounds $R$, pruning rate $p$
\Ensure Winning ticket (sparse mask) $m$

\State Randomly split dataset $D$ into two halves $D^{(1)}$ and $D^{(2)}$ based on class labels
\State Initialize mask $m = \mathbf{1}^{|\theta|}$ (all ones, same size as $\theta$)

\For{$r = 1$ to $R$}
    \State Train model $\mathcal{F}_1(x; m \odot \theta^{(1)})$ on $D^{(1)}$
    \State Train model $\mathcal{F}_2(x; m \odot \theta^{(2)})$ on $D^{(2)}$
    
    \State Generate intermediate masks:
    \State $m^{(1)} = \text{prune\_lowest\_magnitude}(\theta^{(1)}, p\%)$
    \State $m^{(2)} = \text{prune\_lowest\_magnitude}(\theta^{(2)}, p\%)$
    \If{$r < R$}
        \State Rewind unpruned weights in $\theta^{(1)}$ and $\theta^{(2)}$ to their initial values
    \State Update mask: $m = m^{(1)} \cap m^{(2)}$ (element-wise AND operation)

    \EndIf
\EndFor

\State \Return final mask $m$
\end{algorithmic}
\end{algorithm}

This iterative pruning process will continue until a validation accuracy of the entire dataset, $D$, becomes equal to or better than the accuracy original unpruned model. Because of the train-prune-rewind cycle on the overlapping masks/tickets, we name the ticket generated by our method as Cyclic Overlapping Lottery Ticket (COLT). Fig. \ref{fig:colt_illustration} provides a visual illustration for generating COLT. Note that we generate COLT based on an architecture that can assign prediction scores for half of the total classes (either $\mathcal{Y}^{(1)}$ or $\mathcal{Y}^{(2)}$ classes). However, we want to evaluate COLT on the entire dataset ($\mathcal{Y}$ classes), which means the same architecture needs to predict scores for all categories. Therefore, to create our final model, $\mathcal{F}$, we removed the winning tickets' output layer and replaced it with a randomly reinitialized fully connected layer to predict scores for all classes. We then train the tickets until convergence before reporting the final performance on the full dataset. Algorithm \ref{algorithm} summarises the overall process of our proposed method.

We outline some key features and advantages of our pruning approach.
\textit{\textbf{(a)} Utilizing transfer property:} We investigate the `transfer' property of lottery tickets within different partitions of the same dataset. It helps to confidently identify prospective weights to be pruned in earlier pruning rounds. 
\textit{\textbf{(b)} Faster pruning:} 
COLT utilizes a novel approach of overlapping two masks to generate highly sparse neural networks in significantly less time than LTH. By requiring fewer iterations to achieve similar sparsity levels, COLT considerably speeds up the pruning process. Furthermore, by preserving essential weights, COLT maintains a similar or better performance than LTH. Additionally, This heuristic is more efficient than the traditional unstructured pruning heuristic used in LTH.
\textit{\textbf{(c)} Avoiding layer collapse:} Even after completing pruning two times in a single round, our method can avoid layer collapse \cite{r39}. Tickets from the first and second partitions are transferrable and strongly correlate with each other. Thus, intersecting $m^{(1)}$ and $m^{(2)}$ do not discard weights such that a complete layer gets collapsed.
\textit{\textbf{(d)} Introducing Regulirization:} By retaining only high-magnitude weights and zeroing out lower-magnitude ones, COLT implements a form of implicit regularization, reducing the model’s capacity to overfit.

\subsection{Pruning heuristics}\label{pruningheuristics}

Pruning heuristics of our proposed method is motivated by previous work \cite{r3,r5}. We describe different aspects of our pruning heuristics below:

\begin{itemize}

    \item Pruning can either be structured or unstructured. Structured pruning discards weights in clusters, i.e., by removing layers/neurons (weight columns), filters, or channels in CNNs. Unstructured pruning does not follow any specific rule, where weights are pruned in a scattered manner to bring sparsity. Since COLTs are generated by selecting the lower magnitude weights from the entire network, they can prune weights from multiple random layers in one pruning round. Therefore, our strategy follows unstructured pruning.

    \item Based on the strategy used, pruning can be local or global. Local pruning is about pruning a fixed fraction of weights from each layer. In contrast, at each pruning round, global pruning involves pruning a fixed fraction of weights from the entire network allowing each layer to have a different percentage of weights to be discarded. We identify our pruning strategy as global since we treat the whole network at once.
    \item Pruning can be performed either once (one-shot pruning) or iteratively (iterative pruning). Pruning a significant fraction of the weights in one round can be noisy and eliminate important weights. To overcome this problem, we follow an iterative approach, i.e., several iterations of alternating train-prune cycles with a fraction of the weights pruned at once. This generates significantly better-pruned models and winning tickets \cite{r3, r32}. 


    \item Our approach uses a novel heuristic to calculate winning tickets from overlapping weights between two models trained on different segments of a dataset. It makes the winning tickets more robust to class variation. Also, it increases the generalizing capability of the tickets while transferring to another dataset.

    
\end{itemize}

\section{Experiments}\label{experimentsection}

\subsection{Setup}\label{setup}

\begin{table}
\centering
\caption{Model architectures examined in this work. Brackets denote residual connections around layers.}
\label{tab:all_archs}
\begin{tabular}{lcccc} 
\hline
\textbf{Network} & \textbf{Conv-3}                                                        & \textbf{MobileNetV2}                                                                                        & \textbf{ResNet-18}                                                                               & \textbf{VGG-16}                                                                                                                            \\ 
\hline
\textbf{Conv}    & \begin{tabular}[c]{@{}c@{}}64, pool\\128, pool\\256, pool\end{tabular} & \begin{tabular}[c]{@{}c@{}}6$\times$[34,34]\\5$\times$[17,17]\\8$\times$[9,9]\\31$\times$[5,5]\end{tabular} & \begin{tabular}[c]{@{}c@{}}16\\3$\times$[16,16]\\3$\times$[32,32]\\3$\times$[64,64]\end{tabular} & \begin{tabular}[c]{@{}c@{}}2$\times$64, pool\\2$\times$128, pool\\3$\times$256, pool\\3$\times$512, pool\\3$\times$512, pool\end{tabular}  \\ 
\hline
\textbf{Weights} & \begin{tabular}[c]{@{}c@{}}All: 397K\\Conv: 371K\end{tabular}          & \begin{tabular}[c]{@{}c@{}}All: 2.3M\\Conv: 2.2M\end{tabular}                                               & All: 11M                                                                                         & \begin{tabular}[c]{@{}c@{}}All: 136M\\Conv: 14\end{tabular}                                                                                \\ 
\hline
\textbf{P}       & Conv 15\%                                                              & Conv 15\%                                                                                                   & Conv 15\%                                                                                        & \begin{tabular}[c]{@{}c@{}}Conv 15\%\\FC 20\%\end{tabular}                                                                                 \\
\hline
\end{tabular}
\end{table}

\noindent\textbf{Datasets:}
\label{datasetsection}
We perform experiments on three image datasets. 

\noindent\textit{\textbf{(a)} Cifar-10:} Cifar-10 dataset contains 60000 color images (32$\times$32 resolution) belonging to ten classes having 6000 images per class\cite{r25}. 

\noindent\textit{\textbf{(b)} Cifar-100:} Cifar-100 \cite{r25} dataset is similar to Cifar-10 in terms of image resolution and the total number of images. However, it has 100 classes with 600 images per class. 

\noindent\textit{\textbf{(c)} Tiny ImageNet:} Tiny ImageNet dataset \cite{r31} is a scaled-down version of ImageNet introduced in the MicroImagenet classification challenge. It has 100000  color images (64$\times$64 resolution) with 200 classes and 500 images per class. 
\noindent\textit{\textbf{(d)} ImageNet:} The ImageNet \cite{r48} is a large-scale dataset designed for use in visual object recognition software research. Originally introduced in the Large Scale Visual Recognition Challenge (ILSVRC), it contains over 14 million images and spans 1000 classes. Each class in ImageNet typically contains several hundred to over a thousand images, providing a diverse and extensive sample of each category. 
\noindent\textit{\textbf{(e)} Pascal VOC:}\label{pascalvocdataset}
The Pascal VOC (Visual Object Classes)\cite{r45} dataset is a well-known dataset for benchmarking object detection and segmentation tasks. It contains images of various sizes, taken from natural scenes, with each image labeled with objects of one or more of 20 different classes (such as person, car, bird, etc.). The dataset contains 9,963 images and is divided into three sets: training, validation, and testing. The training set is used to train the models, the validation set is used to tune the hyperparameters, and the testing set is used to evaluate the final performance of the models. Each image in the dataset is annotated with ground-truth bounding boxes for the objects in the image and the class label for each object. All the datasets have been systematically partitioned into 
$N$ distinct subsets to facilitate a more specialized analysis, ensuring that each subset contains an equal number of classes. This partitioning strategy deviates from \cite{r5}, which distributed the dataset into parts where all classes were present equally in each partition.




\begin{figure*}[!t]
 	      \centering
            \subfloat{\includegraphics[width=0.33\linewidth]{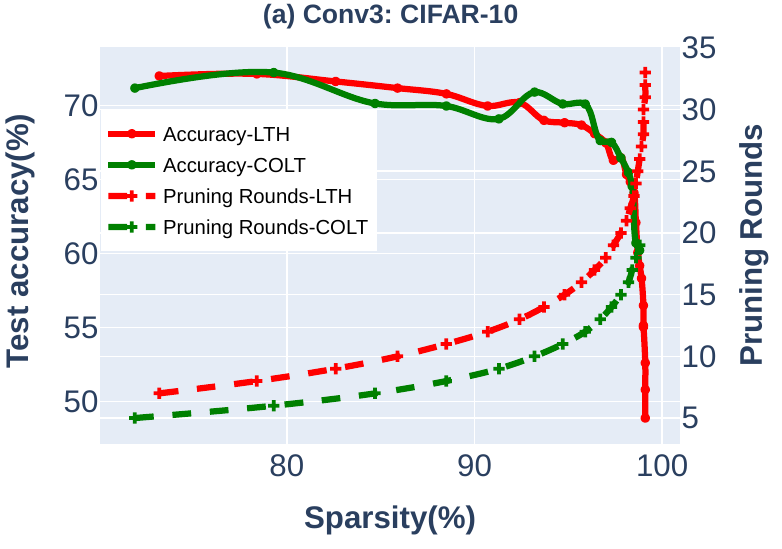}}
            \hfill
            \subfloat{\includegraphics[width=0.33\linewidth]{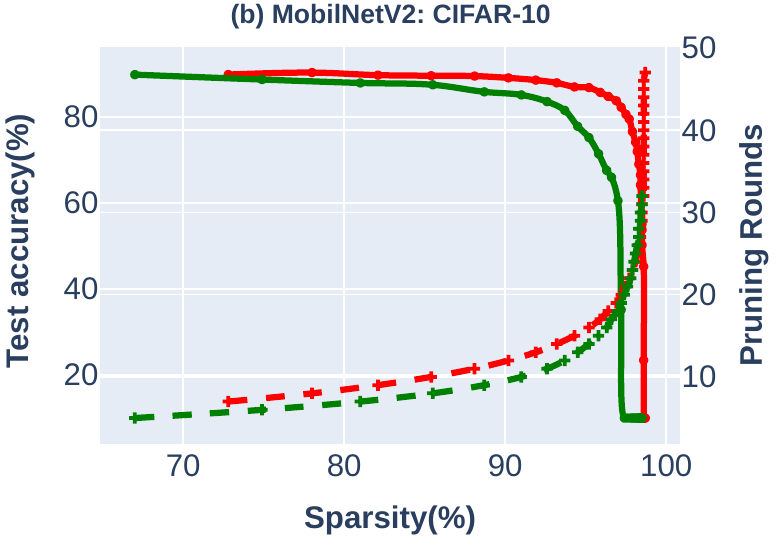}} 
            \hfill
     	\subfloat{\includegraphics[width=0.33\linewidth]{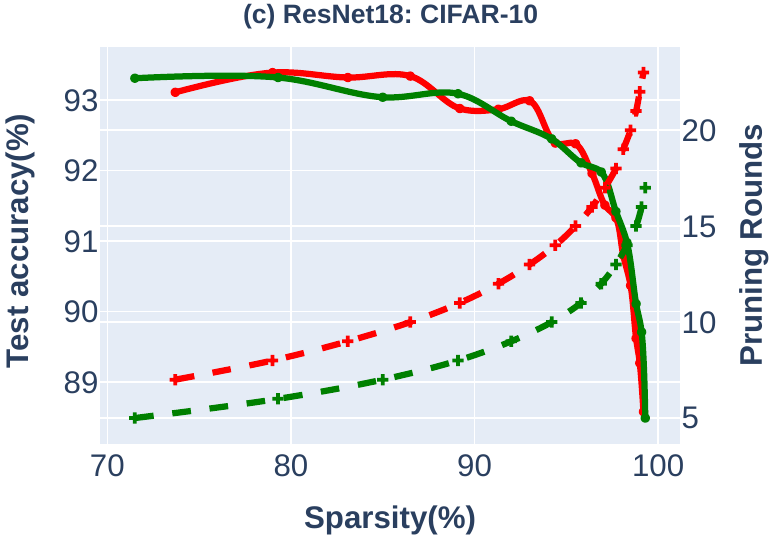}}
            \hfill
     	\subfloat{\includegraphics[width=0.33\linewidth]{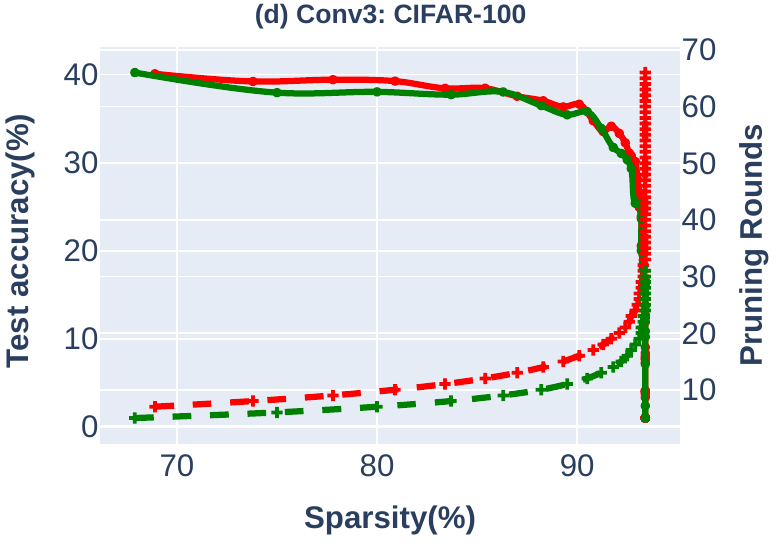}}
     	\hfill
     	\subfloat{\includegraphics[width=0.33\linewidth]{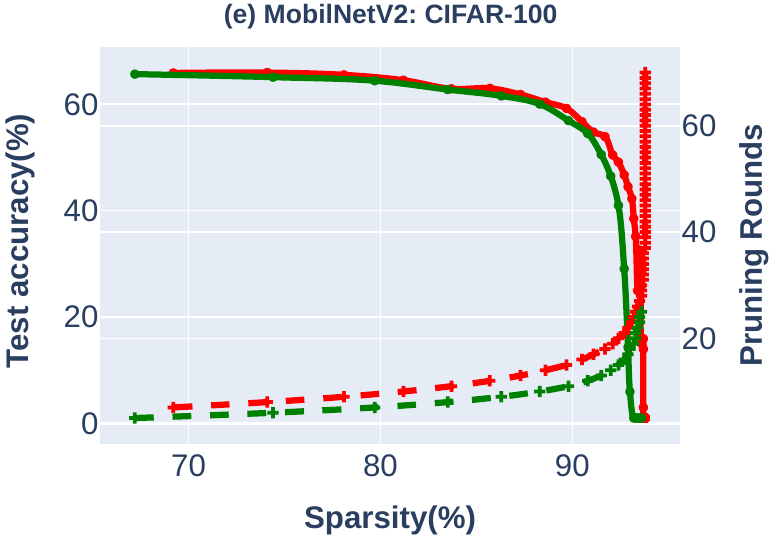}}
     	\hfill
     	\subfloat{\includegraphics[width=0.33\linewidth]{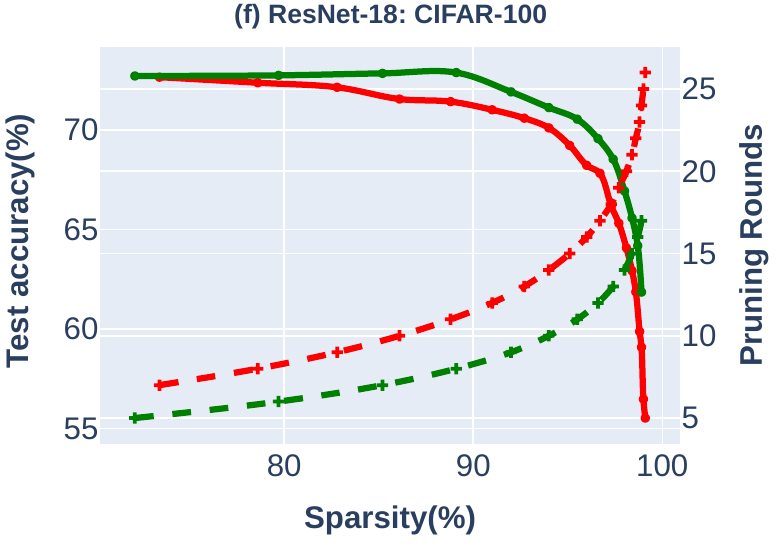}}
     	\hfill
     	\subfloat{\includegraphics[width=0.33\linewidth]{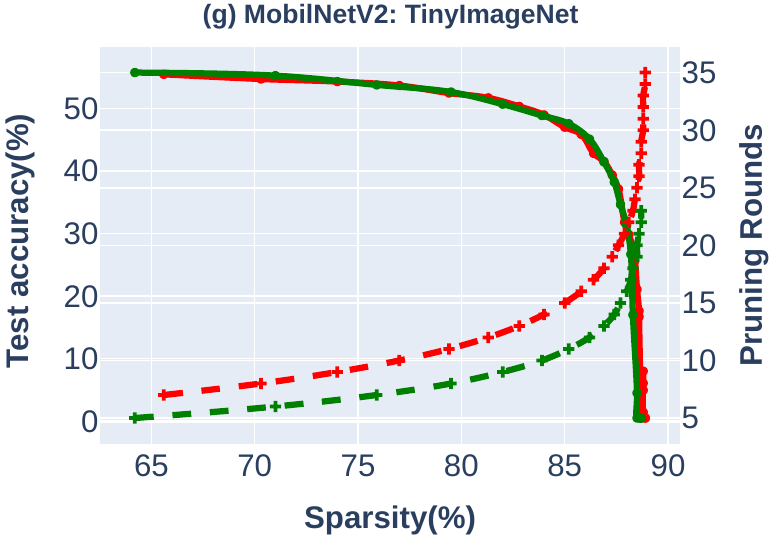}}
     	\subfloat{\includegraphics[width=0.33\linewidth]{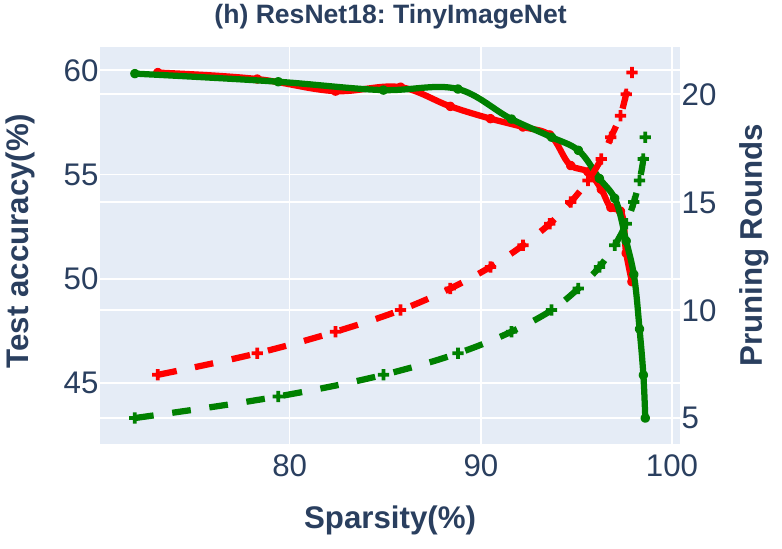}}
      \subfloat{\includegraphics[width=0.33\linewidth]{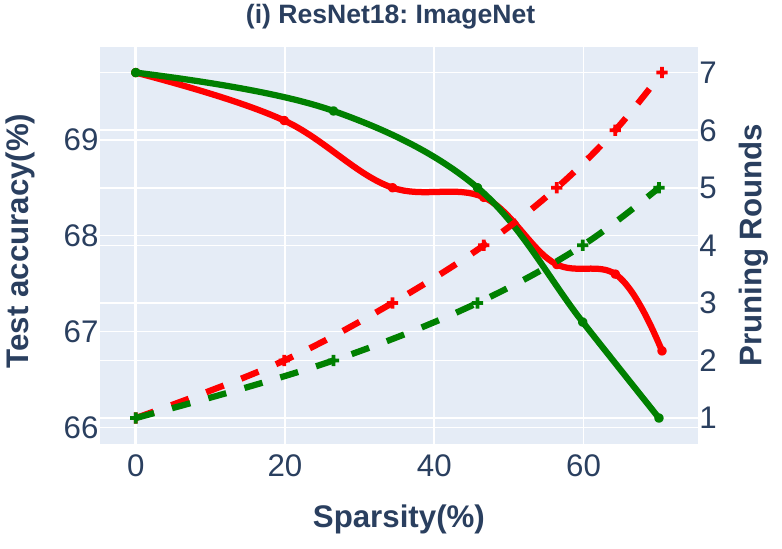}}
        \caption{Performance comparison of LTH and our proposed COLT in terms of accuracy (left y-axis) and pruning rounds (right y-axis) while the sparsity increases (in x-axis). We have used three different datasets (Cifar-10, Cifar-100, TinyImageNet) and three different model architectures (Conv3, MobileNetV2, ResNet18). The \textcolor{red}{red} lines denote LTH, and the \textcolor{green}{green} lines denote COLT. The dotted curves (- - -) represent the pruning rounds taken to reach a desired sparsity, while the solid curves (---) represent the accuracy as the sparsity increases. Observing the solid curves, we can see that the performance of both LTH and COLT starts decreasing when higher sparsities are achieved. However, from dotted curves, we notice that the pruning rounds required by COLT and LTH to reach a particular sparsity are not the same. COLT requires fewer pruning rounds than LTH to achieve the same sparsity while maintaining similar accuracy. Concretely, for any sparsity level, COLT-based pruning costs fewer pruning rounds in comparison to LTH. Therefore, COLT can prune faster than LTH while maintaining decent accuracy.}
 	\label{results}
\end{figure*}

\begin{figure*}[!t]
 	      \centering
            \subfloat{\includegraphics[width=0.33\linewidth]{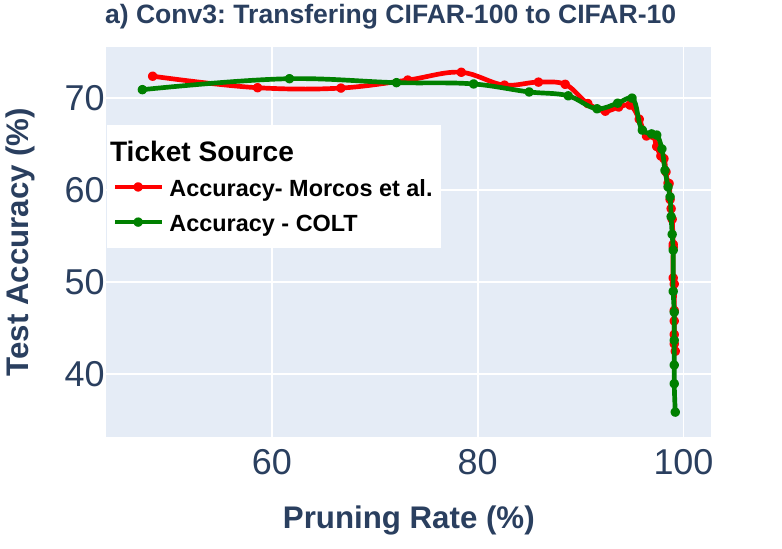}}
            \hfill
            \subfloat{\includegraphics[width=0.33\linewidth]{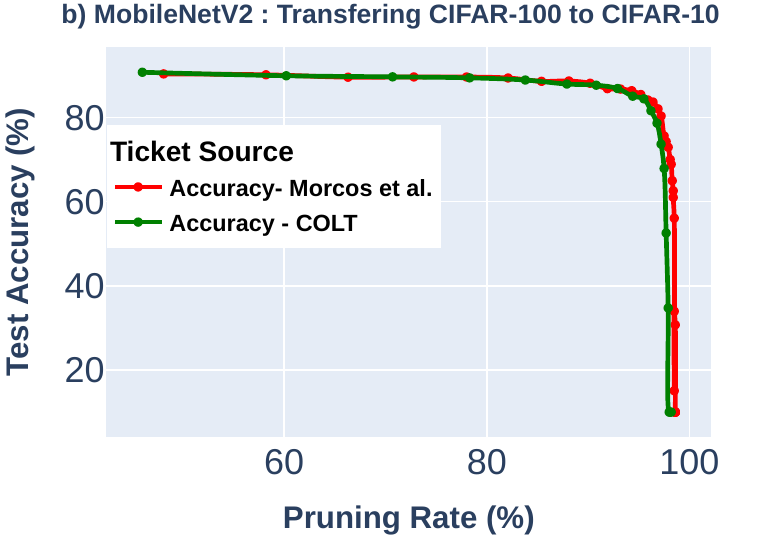}} 
            \hfill
     	\subfloat{\includegraphics[width=0.33\linewidth]{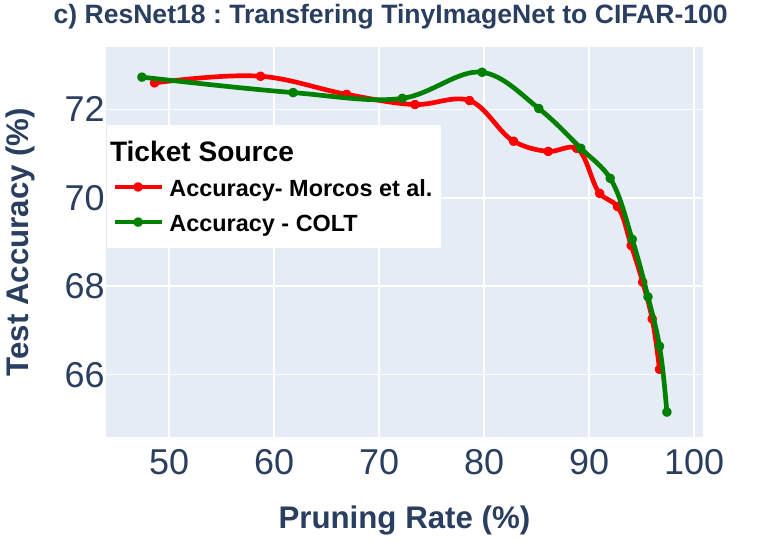}}
            \hfill
     	\subfloat{\includegraphics[width=0.33\linewidth]{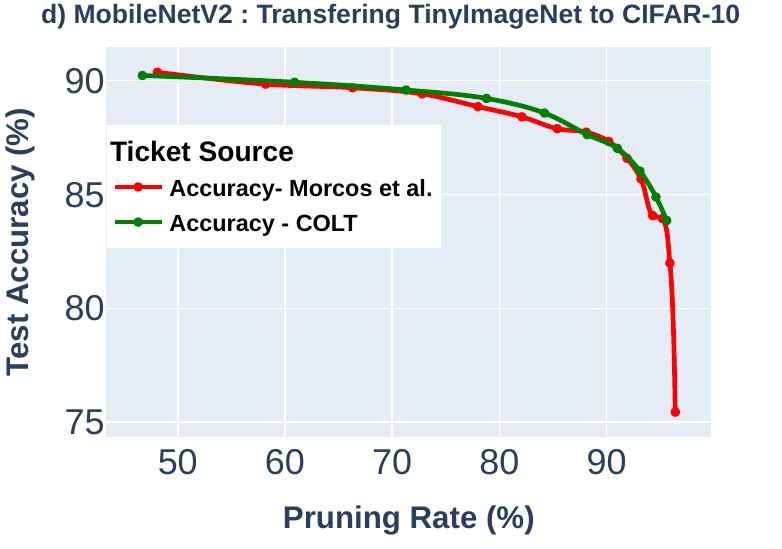}}
     	\hfill
     	\subfloat{\includegraphics[width=0.33\linewidth]{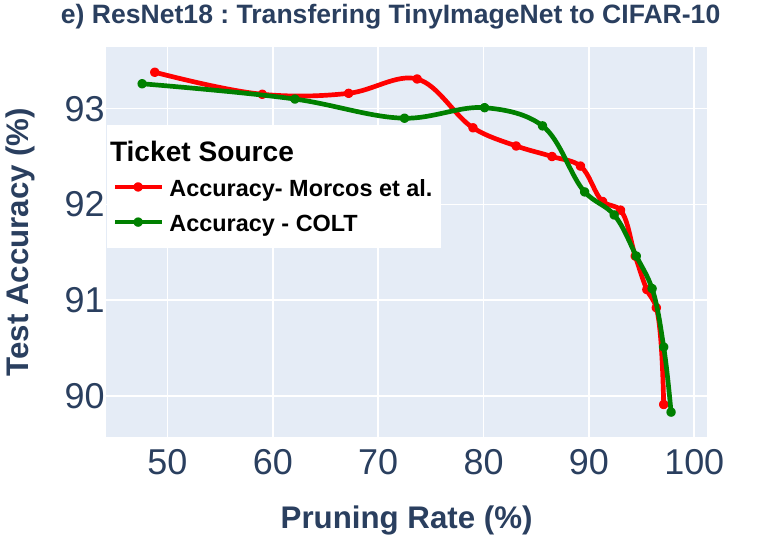}}
     	\hfill
     	\subfloat{\includegraphics[width=0.33\linewidth]{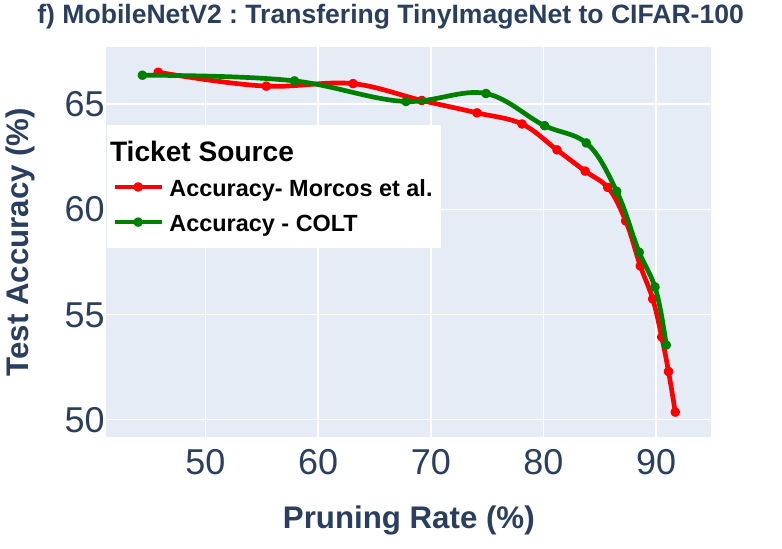}}
     	\hfill
        \caption{Performance comparison of COLT with Morocos et al. \cite{r5} when transferring tickets calculated from one dataset to another. Both methods' performance declines slightly when the tickets reach high sparsities. However, when tickets are transferred from a large dataset (TinyImageNet) to small datasets (Cifar-10, Cifar-100), COLT (\textcolor{green}{green} curve) consistently outperforms Morcos et al. (\textcolor{red}{red} curve) at high sparsities.
        }
 	\label{transfer}
\end{figure*}

\noindent\textbf{Evaluation Process:} To evaluate our models, we have used accuracy, Eq. \ref{accuracy} (in percent), which is a standard evaluation metric for image classification. Additionally, to quantitatively determine the sparsity of our models, we have used prune rate, Eq.  \ref{prune_rate} (in percent) that gives the percentage of parameters that are zero out of all the parameters of a model. Our iterative pruning algorithm runs until a desired $PruneRate$ is reached. 
Each iteration of the pruning algorithm runs for fifty epochs, and the best model (with the lowest validation loss) is selected. 
\begin{equation}
    \label{accuracy}
    Accuracy = \mathit{\frac{Number \: of \: correct \: predictions}{Total \: number \: of \: predictions} \times 100}
\end{equation}

\begin{equation}
    \label{prune_rate}
    Prune Rate  = \mathit{\frac{Number \: of \: zero \: params}{Total \: number \: of \: params} \times 100}
\end{equation}
For object detection tasks, we train 50 epochs similar to the image classification tasks. We calculate the commonly used mean Average Precision (mAP) to evaluate the detection framework.

\noindent\textbf{Model Architecture:}  
We have used three models for conducting experiments: Conv-3, ResNet-18, and MobileNetV2 (see Table \ref{tab:all_archs}). We have used the Conv and ResNet architectures since they were tested in LTH \cite{r3}. We have used the MobileNet architecture because it is a lightweight deep-learning model with few parameters and high classification performance. Conv-3 model architecture is a scaled-down version of VGG \cite{r6} consisting of about 0.37M parameters. It has three convolutional layers, followed by one linear output layer. Besides, we include a batch normalization layer followed by max-pooling after each convolution layer. The ResNet-18 architecture consists of 72 layers with 18 deep residual connection layers \cite{r32} and about 11.1M parameters. MobileNetV2 \cite{mobilenetv2} architecture consists of 53 layers with 19 residual bottleneck layers and about 2.2M parameters. We use Xavier or Glorot weight initialization technique for all the models \cite{r36}. For all three models, the last layer is a fully-connected output layer from the global average pool of the final convolution layer to the number of output classes as done in \cite{r3,r32}. The last convolutional layer was globally average pooled to ensure that we could transfer the weights of a model trained on one dataset to another dataset of a different input shape without bringing any modifications to the model. We trained all three models on Cifar-10 and Cifar-100, whereas Conv-3 is excluded from Tiny ImageNet because it is too shallow to train on Tiny ImageNet.

We also perform COLT-based pruning for object detection tasks based on Faster-RCNN  \cite{r46} and Pascal VOC dataset \cite{r45}. The Faster-RCNN model comprises two main components: a Region Proposal Network (RPN) and a Fast R-CNN network. The RPN generates region proposals by sliding a small network over the convolutional feature map of the input image. The Faster R-CNN network then classifies the proposed regions and refines their bounding boxes. We use VGG16 \cite{r47} as the backbone, a shared feature extraction network between RPN and the detection network containing 13 convolutional layers. The Faster-RCNN network consists of an ROI pooling layer and two fully connected layers for classifying the proposed regions and refining their bounding boxes. This model had 136 million parameters, making it a large and computationally expensive model.



\noindent\textbf{Implementation details:}
After training and pruning, weights must be reset to their initial values at the beginning of training (training iteration 0) to generate winning tickets \cite{r3}. However, Frankle et al. \cite{r3} have found that this only works for shallow models. For deeper models, a learning rate warmup is necessary, along with weight resetting to generate winning tickets. So, our experiments use a learning rate warmup for the first epoch. We globally prune only the convolutional layers at a rate of 0.2 (20\%) for LTH and 0.15 (15\%) for COLT per iteration. Similarly, we do not prune the linear output layers as they comprise only a tiny fraction of the whole network. Each model on cifar10,cifar100, and Tiny Imagenet datasets is trained for 50 epochs with a batch size of 64 and a learning rate of 0.1. The learning rate is annealed by a factor of 5 at 25, 35, and 45 epochs. While training Imagenet with the ResNet-18 model, we trained the model for 90 epochs with a batch size of 512 and a learning rate of 0.2. We have used a learning rate schedular to anneale the learning rate at 30,60 and 80 epochs by a factor of 5. All the models, except Conv-3, are trained using Stochastic Gradient Descent (SGD) with a momentum of 0.9 and weight decay of 0.0005. Conv3 is trained using an Adam optimizer with betas of 0.9 and 0.999 and weight decay of 0.0001.
For the object detection task, we implement the Faster-RCNN model. To train the model, we use the Adam optimizer with a learning rate of 0.0001 and a cosine annealing learning rate scheduler with a minimum learning rate of 1.02e-06. We use a batch size of 4 and train the model for 50 epochs. The input shape is [600,600], and we prune 15\% of the weights of convolutional layers and 20\% of the weights of fully connected layers per pruning round. We implement this work with \textit{PyTorch} framework using a single RTX 3090 GPU.


\subsection{Compared methods}


We compare our work with the three methods discussed as follows: \textbf{(1) Unpruned network:} The unpruned network is the original CNN (Conv-3/Mobilenetv2/Resnet-18) trained until convergence, where no pruning strategy is considered during the learning process. We initialize their weights using the Xavier/Glorot weight initialization technique \cite{r36}. We assume that an unpruned network is generally overparameterized and, therefore, a pruning method is needed to discard unnecessary weights.
\textbf{(2) Lottery Ticket Hypothesis (LTH) \cite{r3}:} To compare with LTH, we also generated LTH tickets. We pruned 20\% weights after each training iteration and then rewound them to their initial state. We then compare their performance with COLT tickets in Sec. \ref{resultssection}. \cite{r3} and \cite{r5} have shown that random tickets with and without masking preserved results in lower performance than winning tickets with equal parameters. As a result, we have omitted using random tickets while evaluating COLT tickets and directly compared COLT tickets with LTH tickets.
\textbf{(3) Morcos et al. \cite{r5}:} Winning tickets are computationally expensive to generate because of the repetitive train-prune-rewind cycle. Morcos et al. have shown that once generated from a dataset, winning tickets can be transferred to another dataset, bypassing the need to find separate winning tickets for each dataset. Concretely, they observed that transferring the winning tickets from one dataset to another can achieve almost similar performance to the winning tickets generated on the same dataset. They have shown that winning tickets generalize well in the natural image domain. Moreover, they have also demonstrated winning tickets generated from large datasets consistently transferred better than those generated using small datasets. 

\noindent\textbf{ COLT vs. other network compression methods: } 
 Although numerous techniques for network compression are available for CNNs, such as knowledge distillation \cite{r28}, quantization \cite{r53}, and structured pruning \cite{r37}, these strategies often necessitate complete \textit{pre-training} of the network or yield models that must undergo \textit{fine-tuning}. This is different from our objective of finding trainable sparse subnetworks from \textit{scratch}. Therefore, in line with previous work \cite{r33,r54}, we do not compare our work with other network compression methods. COLT's primary contribution lies in improving the ticket-finding process itself while maintaining the key advantage of LTH. This focus on initialization-time sparsity and trainability sets lottery ticket methods apart in the broader landscape of network compression.

\subsection{Main Results}\label{resultssection} 
We generate COLT and LTH tickets for ResNet-18 and MobileNetV2 on Cifar-10, Cifar-100, and Tiny ImageNet. For Conv-3, we obtain tickets for Cifar-10 and Cifar-100 only following \cite{r3}. For the weight transferability experiment, we transfer the tickets calculated from Tiny ImageNet to Cifar-10 and Cifar-100. Again, we transfer the tickets from Cifar-100 to Cifar-10. All the experiments generating COLT and LTH tickets are repeated multiple times with a different random initialization, and we reported the best scores.

\noindent\textbf{Results on Cifar-10:} We observe that the accuracy of COLT and LTH are level pegging for ResNet-18. For Conv-3 tickets (Fig.~\ref{results}(a)), we notice a similar picture with both COLT and LTH performing equally. For MobileNetV2 (Fig.~\ref{results}(b)), we see that the performance of COLT and LTH are similar till the pruning rate of 70\%, and then the performance of COLT tickets starts decreasing compared to LTH tickets. For both methods, the performance of the tickets drops steeply when the pruning rate is higher than 95\% for Conv-3, 90\% for MobileNetV2, and 96\% for ResNet-18. For all tickets, COLT consistently requires fewer pruning rounds than LTH to generate a particular sparse ticket. For example, to reach 70\% sparsity, COLT needs 3, 1, and 2 fewer rounds than LTH on Conv-3, MobileNetV2, and ResNet-18, respectively. Similarly, to achieve 90\% sparsity, COLT requires 4, 2, and 3 fewer rounds than LTH on Conv-3, MobileNetV2, and ResNet-18, respectively. Lastly, to reach 98\% sparsity, COLT needs 9, 2, and 5 fewer rounds than LTH on Conv-3, MobileNetV2, and ResNet-18, respectively. Because of the pruning of non-overlapping weights in COLT, it consistently requires fewer rounds than LTH to generate sparse winning tickets. COLT also retains a similar performance to LTH as these non-overlapping weights are unimportant, and excluding those weights does not affect the overall performance much.

\noindent\textbf{Results on Cifar-100:} Observing Fig..~\ref{results}(f), the performance of COLT and LTH tickets is similar until a prune rate of 80\% for ResNet-18. However, after 80\% pruning, the COLT-based tickets perform better than LTH tickets. For MobileNetV2 (Fig.~\ref{results}(e)) and Conv-3 Fig.~\ref{results}(d), the performance of COLT and LTH tickets are almost similar, with LTH slightly outperforming COLT at some prune rates. In addition, the accuracy of the LTH and COLT tickets starts to drop drastically when the pruning rate is greater than 95\% for Resnet-18, 85\% for MobileNetV2, and 90\% for Conv-3. Besides, COLT regularly generates winning tickets in fewer pruning rounds than LTH. To reach 70\% sparsity, COLT requires 3, 2, and 2 fewer rounds than LTH on Conv-3, MobileNetV2, and ResNet-18, respectively. To achieve 90\% sparsity, COLT requires 6, 6, and 3 fewer rounds than LTH on Conv-3, MobileNetV2, and ResNet-18, respectively. Lastly, to reach 98\% sparsity, COLT needs 9, 8, and 6 fewer rounds than LTH on Conv-3, MobileNetV2, and ResNet-18, respectively. This becomes possible due to COLT's aggressive non-matching weights pruning strategy at each round. Moreover, COLT preserves accuracy as the non-matching weights are insignificant and do not hamper the accuracy.

\noindent\textbf{Results on Tiny ImageNet:}
Analyzing Fig.~\ref{results}(g) and \ref{results}(h), we can observe that the performance of COLT and LTH tickets are similar for most of the tickets, with COLT tickets slightly performing better than LTH Tickets at some sparsities. Apart from that, both tickets' accuracy drops significantly from 85\% for ResNet-18 and from 75\% for MobileNetV2. Looking at the pruning rounds, we can see that COLT always generates sparse tickets at fewer iterations than LTH. To reach 70\% sparsity, COLT requires two fewer rounds than LTH on MobileNetV2 and ResNet-18. To reach 90\% sparsity, COLT requires 9 and 3 fewer rounds than LTH on MobileNetV2 and ResNet-18, respectively. Finally, to reach 98\% sparsity, COLT requires 4 and 6 fewer rounds than LTH on MobileNetV2 and ResNet-18, respectively. As COLT prunes a higher rate of redundant weights than LTH at each iteration, it can generate winning tickets in fewer rounds than LTH without compromising accuracy.

\noindent\textbf{Results on ImageNet:}
In an extended analysis on ImageNet using ResNet-18, both COLT and LTH methods maintained near-baseline accuracy after significant model reductions. Initially, no loss in accuracy (69.7\%) was observed without pruning, as expected. With increased sparsity, COLT's accuracy decreased slightly to 69.3\% (at a 26.5\% pruning rate) and further to 66.1\% (at a 70.11\% pruning rate). LTH showed a similar decline, reaching 66.8\% accuracy (at a 70.52\% pruning rate with a slight fluctuation). The efficiency of COLT is evident as it achieves 70\% pruning in just four rounds, two rounds fewer than LTH. This underscores COLT's aggressive yet effective pruning, offering latency and model efficiency advantages. The results imply that on large-scale datasets, eliminating weights (inconsistent across models) trained on different subsets of classes and those with lesser magnitudes has a minimal impact on model performance. This hints at the redundant nature of these weights in maintaining high accuracy levels.

\noindent\textbf{Impact of Non-uniform (Non-IID) Data Partition:}
To create a non-IID data partition, we implemented COLT\_niid, randomly removing 0-30\% samples from each class. This approach introduces data imbalance across classes, as some classes retain all their samples while others lose up to 30\%. The resulting dataset maintains all original classes but with varying sample sizes, creating a non-uniform distribution that challenges the model's ability to learn from irregularly distributed data. 
 We tested both COLT and LTH using ResNet-18 on CIFAR-100, comparing their performance at various pruning rates under both IID and non-IID settings. The results, shown in Table \ref{table:comparison}, indicate that both methods experience some performance degradation under non-IID conditions. At the 89. 1\% pruning rate, LTH shows a decrease of 1.54 percentage points (from 71.42\% to 69.88\%), while COLT experiences a reduction of 2.24 percentage points (from 72.88\% to 70.64\%). Despite this degradation, both methods maintain reasonable performance levels, with COLT showing comparable resilience to LTH in handling data heterogeneity.

\begin{table}
\centering
\caption{Ablation study on multiple partitions (COLT-2, COLT-3 and COLT-4) using ResNet18 architectures.}
\label{tab:partsfour}
\begin{tabular}{cccccc} 
\hline
\multirow{2}{*}{}    & Method  & Acc. $\uparrow$        & Pruning $\uparrow$   & Latency $\downarrow$   & \# of Pruning $\downarrow$  \\
                           &  & (\%)                   & Rate (\%)            & (minutes)              & Rounds                      \\
\hline
\multirow{5}{*}{\rotatebox[origin=c]{90}{CIFAR10}} & Unpruned & \textbf{94.4} & 0 & - & - \\
& LTH & 92.3 & 97.7 & 320 & 17 \\
& COLT-2 & 92.4 & \textbf{97.7} & 276 & 12 \\
& COLT-3 & 92.2 & 97.3 & 225 & 9 \\
& COLT-4 & 89.2 & 97 & \textbf{148} & \textbf{6} \\
\hline
\multirow{5}{*}{\rotatebox[origin=c]{90}{CIFAR100}} & Unpruned & \textbf{73.3} & 0 & - & - \\
& LTH & 66.2 & 97.3 & 479 & 17 \\
& COLT-2 & 68.4 & \textbf{97.4} & 355 & 12 \\
& COLT-3 & 68.0 & 97.3 & 236 & \IH{8} \\
& COLT-4 & 67.7 & 97.2 & \textbf{188} & \textbf{6} \\ 
\hline
\multirow{5}{*}{\rotatebox[origin=c]{90}{TinyImageNet}} & Unpruned & \textbf{59.9} & 0 & - & - \\
& LTH & 53.2 & 97.3 & 2625 & 18 \\
& COLT-2 & 53.9 & \textbf{97.4} & 1756 & 12 \\
& COLT-3 & 53.4 & 97.6 & 1266 & 8 \\
& COLT-4 & 53.7 & 97 & \textbf{930} & \textbf{6} \\
\hline
\multirow{4}{*}{\rotatebox[origin=c]{90}{ImageNet}} & Unpruned & \textbf{69.7} & 0 & - & - \\
& LTH & 66.8 & 70.5 & 10320 & 6 \\
& COLT-2 & 66.1 & 70.1 & 6960 & 4 \\
& COLT-4 & 55.7 & \textbf{74.1} & \textbf{5760} & \textbf{3} \\

\hline
\end{tabular}
\end{table}

\begin{table}[!t]
\centering
\caption{ 
Performance comparison of LTH and COLT under IID and non-IID data distributions on CIFAR-100 using ResNet-18. Values in parentheses show the accuracy difference between IID and non-IID scenarios.} 
\begin{tabular}{ccccc}
\toprule
\multirow{2}{*}{\textbf{Pruning Rate (\%)}} & \multicolumn{2}{c}{\textbf{LTH}} & \multicolumn{2}{c}{\textbf{COLT}} \\
\cmidrule(lr){2-3} \cmidrule(lr){4-5}
& IID & non-IID & IID & non-IID \\
\midrule
85.2 & 71.55 & 70.39  (-1.16)  & 72.84 & 71.33  (-1.51)  \\
89.1 & 71.42 & 69.88  (-1.54)  & 72.88 & 70.64  (-2.24)  \\
92.1 & 70.59 & 69.28  (-1.31)  & 71.91 & 70.30 \ (-1.61)  \\
\bottomrule
\end{tabular}
\label{table:comparison}
\end{table}

\begin{table*}[!t]
\centering
\caption{State-of-the-art results describing the performance comparison of tickets generated by the original unpruned network, LTH, and COLT. The tickets were generated and evaluated on TinyImageNet and then transferred to Cifar-10 and Cifar-100. We compare sparse tickets from three methods on ResNet-18 and MobileNetV2 architectures. $\uparrow$ ($\downarrow$) means higher (lower) is better. ``-"  denotes results that are not applicable there.}
\label{tab:overallresult}

\resizebox{\linewidth}{!}{%
\begin{tabular}{ccccc} 
\hline
\multicolumn{5}{c}{Model: ResNet18 and Dataset: Tiny ImageNet}                                                      \\ 
\hline
\multirow{2}{*}{Method} & Acc. $\uparrow$ & Pruning $\uparrow$ & Latency $\downarrow$ & No of Pruning $\downarrow$  \\
                        & (\%)            & Rate (\%)          & (minutes)            & Rounds                      \\ 
\hline
Unpruned                & \textbf{59.9}            & 0                  & -                    & -                           \\
LTH                     & 53.2            & 97.3               & 2625                 & 18                          \\
Ours (COLT)             & 53.9            & \textbf{97.2}      & \textbf{1756}        & \textbf{12}                 \\ 
\hline\hline
\multicolumn{5}{c}{Tickets transformed from TinyImageNet to Cifar10}                                                \\ 
\hline
Unpruned                & \textbf{93.4}            & 0                  & -                    & -                           \\
Morcos et al. \cite{r5}           & 89.9            & 97.1               & -                    & -                           \\
Ours (COLT)             & 90.5            & \textbf{97.1}      & -                    & -                           \\ 
\hline
\multicolumn{5}{c}{Tickets transformed from TinyImageNet to Cifar100}                                               \\ 
\hline
Unpruned                & \textbf{73.2}            & 0                  & -                    & -                           \\
Morcos et al. \cite{r5}           & 64.8            & 97.3               & -                    & -                           \\
Ours (COLT)             & 65.2            & \textbf{97.4}      & -                    & -                           \\
\hline
\end{tabular}%
\begin{tabular}{lcccc} 
\cline{2-5}
 & \multicolumn{4}{c}{Model: MobileNetV2 and Dataset: Tiny ImageNet}                         \\ 
\cline{2-5}
 & Acc. $\uparrow$ & Pruning $\uparrow$ & Latency $\downarrow$ & \# of Pruning $\downarrow$  \\
 & (\%)            & Rate (\%)          & (minutes)            & Rounds                      \\ 
\cline{2-5}
 & \textbf{56.1}            & 0                  & -                    & -                           \\
 & 54.7            & 70.3               & 856                  & 7                           \\
 & 55.2   & \textbf{71.0}      & \textbf{354}                  & \textbf{5}   \\ 
\cmidrule{2-5}\morecmidrules\cmidrule{2-5}
 & \multicolumn{4}{c}{Tickets transformed from TinyImageNet to Cifar10}                      \\ 
\cline{2-5}
 & \textbf{90.6}            & 0                  & -                    & -                           \\
 & 88.9            & 78.0               & -                    & -                           \\
 & 89.2   & \textbf{78.8}      & -                    & -                           \\ 
\cline{2-5}
 & \multicolumn{4}{c}{Tickets transformed from TinyImageNet to Cifar100}                     \\ 
\cline{2-5}
 & \textbf{65.8}            & 0                  & -                    & -                           \\
 & 64.1            & 74.1               & -                    & -                           \\
 & 65.5   & \textbf{74.9}      & -                    & -                           \\
\cline{2-5}
\end{tabular}
}

\end{table*}

\subsection{Multiple Partitions}\label{multiplepartitions}

In this section, we evaluate COLT's performance with multiple dataset partitions, comparing COLT-2, COLT-3, and COLT-4, representing the method for 2, 3, and 4 partitions, respectively, against the widely used LTH method using ResNet18 models on multiple datasets. Detailed results are shown in Table-\ref{tab:partsfour}. The results indicate that the partition-based pruning approach works differently for datasets of different sizes and complexities. COLT-3 (partitioning three parts of the dataset) has shown a balanced performance with minor accuracy losses and good pruning rates across CIFAR-10 (92.2\% accuracy, 97.3\% pruning), CIFAR-100 (68.0\%, 97.3\%), and Tiny ImageNet (53.4\%, 97.6\%). COLT-4, however, shows reduced accuracy on CIFAR-10 (88.7\%) compared to LTH (91.3\%) due to the small size of CIFAR-10, which makes it hard to maintain informative weights with increased partitioning. Conversely, for CIFAR-100, COLT-4's accuracy (67.7\%) is slightly lower than COLT-3's(68.0) but still above LTH's (66.2\%). For Tiny ImageNet, COLT-4 (53.7\%) slightly outperforms LTH (53.2\%), indicating more complex datasets can preserve essential weights even with more partitions. Suggesting larger datasets can better handle more partitions. 
On ImageNet, the most challenging dataset, aggressive pruning with COLT-4 leads to a significant accuracy drop (55.7\%) despite a high pruning rate (74.01\%) and the lowest latency (5760 minutes), implying a loss of critical weights. It is likely due to the loss of weights critical for underrepresented features within the dataset~\cite{r49}. COLT-2, on the other hand, achieves a closer accuracy to LTH (66.1\% vs. 66.8\%) with a better pruning rate (70.11\%) and lower latency (6960 minutes), showing that moderate pruning strategies like COLT-2 can retain important features on complex datasets, striking a balance between model size, efficiency, and performance.

\subsection{Transferring LTH \& COLT Tickets Across Datasets}
The iterative process of ticket generation can be costly. To mitigate this problem, Morcos et al. \cite{r5} showed that winning tickets (pruned subnetwork) generated on larger, more complex datasets could generalize substantially better than those generated on small datasets. In other words, we do not need to create tickets for each dataset. Tickets generated from one larger dataset can be transferred/employed again to train the same network for a smaller dataset. For this reason, we perform experiments to transfer the COLT and LTH tickets of Tiny ImageNet to Cifar-10 and Cifar-100. Since we are using datasets with different input shapes, there will be a weight mismatch when transferring weights of the linear layer. To overcome this problem, we have globally average pooled the output from the last convolutional layer before passing it through the dense output layer. \cite{r5} further showed that just increasing the number of classes while keeping the dataset size fixed can significantly improve the generalization capability of winning tickets. To verify this point, in our work, we have transferred the COLT and LTH tickets of Cifar-100 to Cifar-10 since they are the same size, but Cifar-100 has more classes than Cifar-10. 

Observing Fig.~\ref{transfer}, we can see that the accuracy slowly deteriorates as COLT and Morcos et al. reach high sparsities. However, COLT consistently performs better than Morcos et al. at high sparsities when TinyImageNet tickets are transferred to Cifar-10 \& Cifar-100. Additionally, both methods are on par and work well when Cifar-100 tickets are transferred to Cifar-10 for all sparsities. Therefore, we claim that COLT tickets have a high generalization capacity and transferability to other datasets of a similar domain.

One can compare the performance of regular tickets (COLT) with transferred tickets by analyzing Fig. \ref{results} and \ref{transfer}. We observe that the performance trend for regular and transferred tickets is similar across architectures and datasets. If a regular ticket performs well on a dataset, the transferred ticket performs well, and vice versa. The bottom line is that we can generate COLT tickets from one dataset and then use the same ticket to train another dataset of a similar domain without compromising network performance.

\subsection{State-of-the-art comparison} 
Table \ref{tab:overallresult} compares the results of unpruned network, COLT, and LTH methods regarding no. of pruning rounds, accuracy, and latency. For the transferability experiment, tickets were generated using TinyImageNet and transferred to Cifar10 and Cifar100. 
Echoing \cite{r5}, transferring tickets from larger (TinyImageNet) to smaller datasets (Cifar10) improves performance. Our studies on ResNet-18 and MobileNetV2 support this. Notably, ResNet-18 maintains accuracy with 97.2\% of parameters pruned, showing that just 2.8\% of weights are needed for satisfactory accuracy. 
COLT outperforms LTH in accuracy at high sparsity with fewer pruning rounds and less training time. Yet, it falls short of the unpruned network's accuracy by 5-6\%, likely because pruning 97\% of weights might remove some crucial ones, leading to accuracy drops. 
In addition, there is a high risk of layer collapse for unstructured at high sparsity. Tanaka et al.\cite{r39} state that iterative magnitude pruning can avert this by enforcing a conservation law through gradient descent, which raises the magnitude scores for larger layers throughout the pruning process. In this regard, COLT performs comparatively better than LTH, even if it receives fewer iterations to prune at high sparsity. Therefore, we claim that COLT can retain ``good weights" even after pruning at high sparsities. On the other hand,  MobileNet has nine million fewer weights than ResNet18. As a result, unlike an unpruned network, neither LTH nor COLT achieves par accuracy at high sparsity. However, the unpruned network and COLT (with 71\% pruning) hold similar accuracy. Likewise, in ResNet18, we generate tickets faster with fewer pruning rounds than LTH. In this case, LTH requires eight pruning rounds and 856 minutes for a 70.3\% sparse ticket with 54.7\% accuracy on TinyImageNet. In contrast, COLT needs a mere five rounds and 354 minutes for a 71\% sparse ticket with 55.2\% accuracy. For MobileNetV2 and ResNet18, COLT is significantly faster than LTH, with fewer rounds and quicker gradient computation. COLT's tickets are sparser. Hence, fewer gradients are needed, allowing for much faster ticket generation than LTH.
    
A similar trend is observed when the tickets are transferred to Cifar-10 and Cifar-100 for medium sparsity tickets (prune rate around 80\%). Particularly for MobileNetV2, both COLT and Morcos et al. \cite{r5} have accuracies close to the original unpruned network. It implies that we can adopt a network that has already been pruned on one dataset and transfer it to another with an accuracy comparable to an unpruned network. Moreover, we do not need additional resources to prune the network for each individual dataset. In contrast, for high-sparsity tickets, the accuracy decreases by about 3\% when transferred to Cifar-10 and by about 8\% when transferred to Cifar-100. However, COLT appears to become more generic than Morcos et al. \cite{r5}. This is because COLT retains the overlapping weights of two networks trained on two one-half splits of a dataset, making the COLT tickets more robust to image variations.

\begin{table*}
\centering
\caption{Per-class mAP of PASCAL VOC dataset after 68\% of pruning.}
\label{tab:classwise}
\scalebox{.9}{
\begin{tabular}{ccccccccccccccccccccc} 
\hline
& \rotatebox{90}{aero.} & \rotatebox{90}{bicycle} & \rotatebox{90}{bird} & \rotatebox{90}{boat} & \rotatebox{90}{bottle} & \rotatebox{90}{bus} & \rotatebox{90}{car} & \rotatebox{90}{cat} & \rotatebox{90}{chair} & \rotatebox{90}{cow} & \rotatebox{90}{table} & \rotatebox{90}{dog} & \rotatebox{90}{horse} & \rotatebox{90}{mbike} & \rotatebox{90}{person} & \rotatebox{90}{plant} & \rotatebox{90}{sheep} & \rotatebox{90}{sofa} & \rotatebox{90}{train} & \rotatebox{90}{tv} \\ 
\hline
Unpruned & \textbf{64.2} & 77.4 & 53.2 & 56.5 & 33.7 & \textbf{77.8} & 82.1 & 69.4 & 46.4 & 56.8 & 55.3 & 67.3 & \textbf{82.4} & 82.1 & 76.3 & \textbf{36.9} & 49.8 & \textbf{54.7} & \textbf{79.7} & 65.3 \\ 
LTH & 63.4 & \textbf{79.1} & 51.4 & 56.7 & 32.4 & 77.7 & \textbf{82.1} & \textbf{71.3} & \textbf{48.1} & 60.4 & \textbf{55.7} & \textbf{68.6} & 81.4 & 81.5 & 77.3 & 31.3 & 46.5 & 54.5 & 78.7 & \textbf{68.4} \\ 
COLT & 63.5 & 77.3 & \textbf{54.1} & \textbf{60.5} & \textbf{34.9} & 77.4 & 81.5 & 69.9 & 44.7 & \textbf{61.4} & 55.3 & 67.1 & 82.2 & \textbf{83.1} & \textbf{77.4} & 35.9 & \textbf{52.7} & 52.4 & 77.7 & 67.8 \\ 
\hline
\end{tabular}}
\end{table*}

\begin{table}\centering
\caption{Pruning the Faster-RCNN model while performing object detection tasks on the PASCAL VOC dataset.}
\label{tab:pascalvoctable}
\begin{tabular}{ccccc} 
\hline
\multirow{2}{*}{Method} & mAP~$\uparrow$ & Pruning $\uparrow$ & Latency $\downarrow$ & No of pruning $\downarrow$  \\
                        & (\%)           & Rate (\%)          & (hours)            & Rounds                      \\ 
\hline
Unpruned                & \textbf{63.4}     & 0                  & -                    & -                           \\
LTH                     & 60.8              & 36.0                 & 26.0                    & 2                           \\
COLT                    & 60.6             & \textbf{32.2}                 & \textbf{14.3}           & \textbf{1}                  \\ 
\hline
LTH                     & 61.6              & 48.8                 & 39.3                    & 3                           \\
COLT                    &\textbf{62.5}              & \textbf{53.5}                  & \textbf{28.8}           & \textbf{2}                  \\ 
\hline
LTH                     & 63.4 
& 67.0                 &  65.5                  & 5                           \\
COLT                    & \textbf{63.8}              & \textbf{68.6}                 & \textbf{43.1}                    & \textbf{3}                  \\ 
\hline
LTH                     &  \textbf{63.7}             & 79.0                 & 91.7                    &  7                          \\
COLT                    & 63.3              & \textbf{79.1}                  & \textbf{58.0}                     & \textbf{4}                           \\ \hline
\end{tabular}
\end{table}

\subsection{Beyound classification}\label{pascalvocresults}
In addition to image classification tasks, we also evaluated the effectiveness of COLT on the Pascal VOC object detection task using the Faster-RCNN model with the VGG backbone. This task is more complex than image classification, as it involves identifying the object and its location in the image. The model is relatively large, with a total of 113 million parameters. Among them, 13 million parameters are attributed to the CNN layers subject to pruning to reduce computational resources. In Table \ref{tab:pascalvoctable}, we compare the performance of COLT with LTH on the Pascal VOC dataset and find that COLT outperforms LTH in terms of pruning rounds, latency, and accuracy. We also report per-class mAP in Table \ref{tab:classwise} to show that after 68\% pruning, individual class mAPs are similar to the unpruned network. Our results demonstrate that COLT and LTH can achieve similar accuracy as the Unpruned network, with a maximum mAP of 63.8\% and 63.7\%, respectively, while pruning up to 79\% of the network. This indicates that our proposed method can achieve a significant compression level with negligible accuracy loss. When comparing COLT with LTH, we observe that COLT achieves similar or better accuracy than LTH, requiring significantly fewer pruning rounds and less latency. 

\begin{figure}[!t]
	\centering
    {\includegraphics[width=\linewidth]{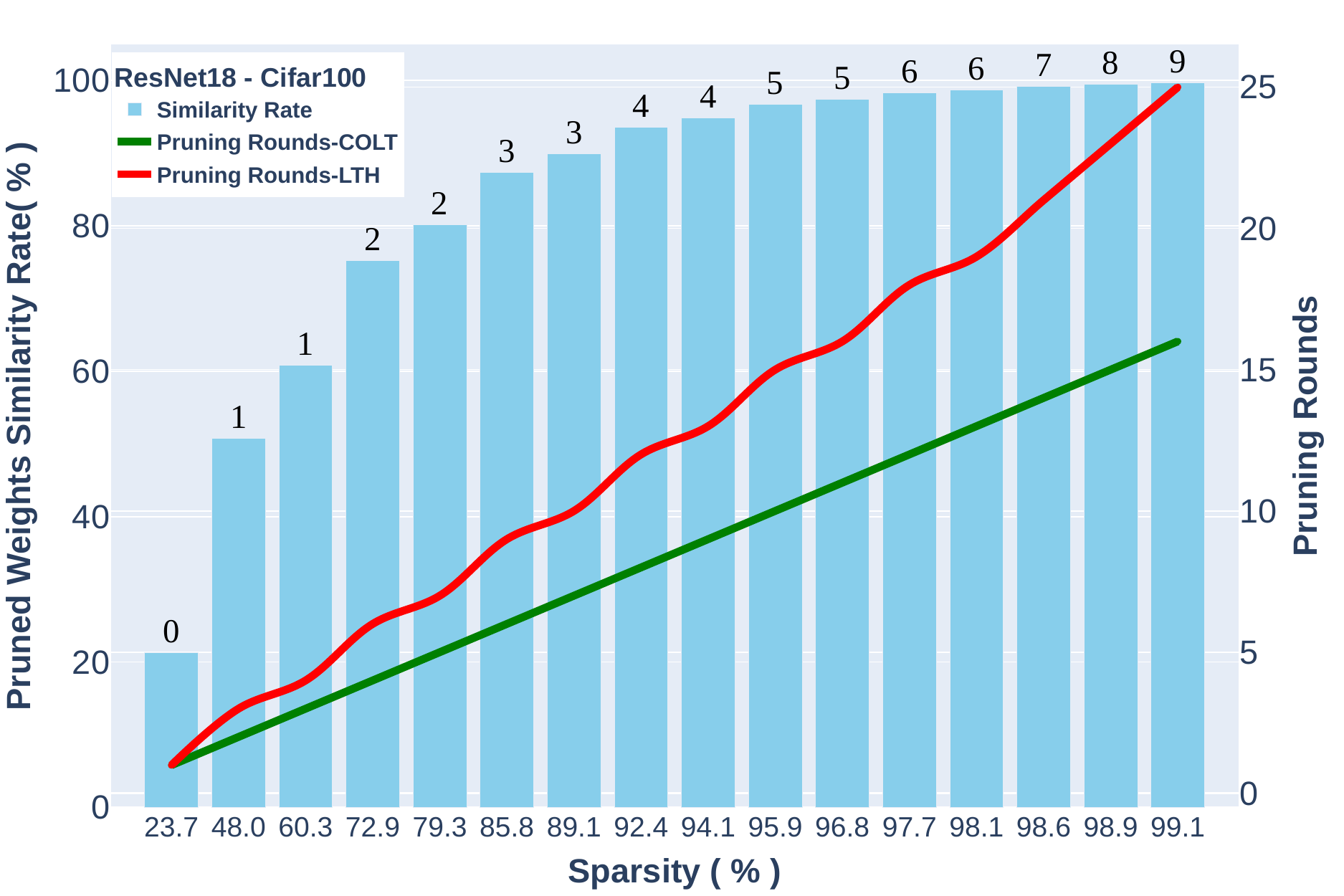}}   
	\caption{The similarity (blue bars) of pruned weights between sparse tickets calculated by COLT and LTH method. With the increase of sparsity, both kinds of methods prune similar weights. To achieve this, COLT requires fewer training rounds than LTH (\textcolor{red}{red} vs. \textcolor{green}{green} curve).}
	\label{similarity}
\end{figure}

\subsection{Discussion}\label{discussion}

\noindent\textbf{Weight Similarity Analysis:}
Fig.~\ref{similarity} depicts a comparison between LTH and COLT in terms of pruned weights similarity and pruning rounds for the ResNet-18 model on the Cifar-100 dataset. The \textit{X-axis} represents the sparsities in percentage. The \textit{Y-axis on the left} represents the pruning rate similarity of unpruned weights, and the \textit{Y-axis on the right} represents the pruning rounds taken to generate a ticket of the same sparsity. The blue bars represent the similarity in the pruned weights between LTH and COLT tickets. The similarity is calculated at the beginning of each pruning round. COLT and LTH tickets are initialized with the same weights. The similarity rate is the number of common pruned parameters between COLT and LTH divided by the total no. of initial parameters. Here, `common' implies that LTH and COLT pruned the same weights. The red and green line joining the bars represents the pruning rounds required by COLT and LTH, respectively. The numbers on the top of the bars depict the additional pruning rounds required by LTH compared to COLT. We find substantially fewer identical remaining weights between COLT and LTH during the early stages. However, as the prune rate increases, the similarity between the remaining weights of LTH and COLT also increases. For example, at the 28\% pruning rate, the similarity of remaining weights is 20\%. Subsequently, the similarity of remaining weights jumps to 50\% when the sparsity of tickets reaches around 48\%. It implies that identical weights between COLT and LTH start getting pruned while the weight matrix becomes sparse. In other words, LTH and COLT essentially prune similar weights to generate a sparse winning ticket. However, COLT (solid red line) requires fewer pruning rounds than LTH (solid green line) to create a ticket of the same sparsity. For example, to reach a sparsity of about 89\%, LTH requires three more pruning rounds than COLT, making the LTH process time-consuming.

\noindent\textbf{Performance Analysis:}\label{analysis} Analyzing all results, we can see that COLT tickets from large models like ResNet-18 are comparable or sometimes superior to LTH tickets across dataset sizes and maintain high accuracy when transferred to different datasets, even at high sparsities. On the other hand, MobileNetV2 COLT tickets tend to underperform on small datasets such as Cifar-10 and Cifar-100, shown in Fig. \ref{results} (b), (c), and (e) relative to LTH tickets. The reason is possibly due to reduced feature diversity after splitting small datasets into smaller partitions, which may lead to difficulty in identifying essential features during pruning. However, their accuracy is comparable when created on larger datasets like Tiny ImageNet. When transferred from Tiny ImageNet to smaller datasets, MobileNetV2 COLT tickets match or exceed LTH performance. Conv-3 COLT tickets also maintain LTH-level performance across datasets, including when Cifar-100 tickets are applied to Cifar-10. It follows that dividing a dataset into $N$ parts results in a smaller dataset, which is subsequently trained using a model with the same number of weights as the full dataset. Consequently, the model may utilize its full potential for the smaller dataset to determine the optimal features and weights for the training samples. Aggressive pruning techniques such as COLT-2, COLT-3, or COLT-4 thus can preserve the best weights to hold accuracy as, or occasionally better than, LTH across the models and datasets.

In object detection, COLT models perform competitively with LTH models at various sparsity levels, with similar mAP scores, underscoring COLT's pruning effectiveness. This indicates COLT's potential as a viable alternative to LTH for pruning, capable of generating sparser models without performance loss and possibly offering simpler and less costly pruning.

\noindent \textbf{ 
Addressing Overfitting:} 
 While neural networks inherently face overfitting challenges, COLT's iterative pruning approach actually serves as a natural regularization mechanism that helps prevent overfitting rather than causing it \cite{r50}. By progressively removing redundant connections while maintaining only the most essential pathways, COLT effectively reduces the model's capacity to memorize training data. Specifically, COLT employs three key mechanisms that contribute to the prevention of overfitting: \textbf{(1)} the retention of only high-magnitude weights while zeroing out lower-magnitude ones serves as implicit regularization \cite{r51}, \textbf{(2)} the overlapping weight selection process across multiple data partitions ensures that only consistently important weights are retained, and \textbf{(3)} the resulting sparse architecture inherently limits the model's capacity to overfit \cite{r52}. Our empirical results support this claim. The performance gap between training and validation accuracy remains stable or even decreases as pruning progresses, particularly at higher sparsity levels (85-92\%). Robust performance under non-IID conditions (see Table \ref{table:comparison}) further demonstrates that COLT-pruned networks learn generalizable features rather than overfitting to specific data distributions.

\section{Conclusion}
In this paper, we proposed a new set of lottery tickets termed COLT to prune deep neural networks. Like tickets from LTH, COLTs can achieve high sparsity and weights' transferability across datasets without compromising accuracy. We compute COLTs by leveraging tickets obtained from two/four class-wise partitions of a dataset. We notice that the performance of these tickets is on par with the LTH-generated lottery tickets in terms of accuracy. Moreover, COLTs are generated at fewer iterations than LTH tickets. We validate our claim using both object recognition and detection tasks. In experiments, we have also demonstrated that COLTs generated on a large dataset (Tiny ImageNet) can be transferred to a small dataset (Cifar-10, Cifar-100) without compromising performance, showing these tickets' generalizing capability. Future works in this line of investigation may consider COLT's sparsity and transferability on deeper architectures (DenseNet/ShuffleNet) and vast datasets (YouTube-BoundingBoxes/MSCOCO). Moreover, one can explore different problem setups (like object segmentation or image captioning) in the context of network pruning.

\bibliographystyle{IEEEtran}
\bibliography{ref2}

\end{document}